\documentclass[10pt,twocolumn,letterpaper]{article}

\usepackage[pagenumbers]{cvpr} %

\usepackage{graphicx}
\usepackage{amsmath}
\usepackage{amssymb}
\usepackage{bm}
\usepackage{booktabs}

\usepackage{multirow}
\usepackage{caption}
\usepackage{subcaption}
\usepackage{enumitem}
\usepackage{xcolor}
\usepackage[export]{adjustbox}
\usepackage{multirow}
\usepackage[accsupp]{axessibility}

\newcommand{\std}[1]{}

\usepackage[pagebackref,breaklinks,colorlinks]{hyperref}

\usepackage[capitalize]{cleveref}
\crefname{section}{Sec.}{Secs.}
\Crefname{section}{Section}{Sections}
\Crefname{table}{Table}{Tables}
\crefname{table}{Tab.}{Tabs.}

\def\cvprPaperID{1779} %
\def\confName{CVPR}
\def\confYear{2023}

\begin{document}

\title{LayoutDM: Discrete Diffusion Model for Controllable Layout Generation}

\author{Naoto~Inoue$^{1}$ \qquad Kotaro~Kikuchi$^{1}$ \qquad Edgar~Simo-Serra$^{2}$ \qquad Mayu~Otani$^{1}$ \qquad Kota~Yamaguchi$^{1}$\\
$^{1}$CyberAgent, Japan \qquad $^{2}$Waseda University, Japan\\
{\tt\small \{inoue\_naoto,~kikuchi\_kotaro\_xa\}@cyberagent.co.jp~ess@waseda.jp} \\
{\tt\small \{otani\_mayu,~yamaguchi\_kota\}@cyberagent.co.jp}
}
\maketitle

\begin{abstract}
Controllable layout generation aims at synthesizing plausible arrangement of element bounding boxes with optional constraints, such as type or position of a specific element.
In this work, we try to solve a broad range of layout generation tasks in a single model that is based on discrete state-space diffusion models.
Our model, named \emph{LayoutDM}, naturally handles the structured layout data in the discrete representation and learns to progressively infer a noiseless layout from the initial input, where we model the layout corruption process by modality-wise discrete diffusion.
For conditional generation, we propose to inject layout constraints in the form of masking or logit adjustment during inference.
We show in the experiments that our LayoutDM successfully generates high-quality layouts and outperforms both task-specific and task-agnostic baselines on several layout tasks.\footnote{Please find the code and models at: \newline \url{https://cyberagentailab.github.io/layout-dm}.}

\end{abstract}

\section{Introduction}
Graphic layouts play a critical role in visual communication.
Automatically creating a visually pleasing layout has tremendous application benefits that range from authoring of printed media~\cite{zhong2019publaynet} to designing application user interface~\cite{deka2017rico}, and there has been a growing research interest in the community.
The task of layout generation considers the arrangement of elements, where each element has a tuple of attributes, such as category, position, or size, and depending on the task setup, there could be optional control inputs that specify part of the elements or attributes.
Due to the structured nature of layout data, it is crucial to consider relationships between elements in a generation.
For this reason, current generation approaches either build an autoregressive model~\cite{gupta2021layout,arroyo2021variational} or develop a dedicated inference strategy to explicitly consider relationships~\cite{lee2020neural,Kikuchi2021,kong2022blt}.

\begin{figure}[t]
    \centering
    \includegraphics[width=\hsize]{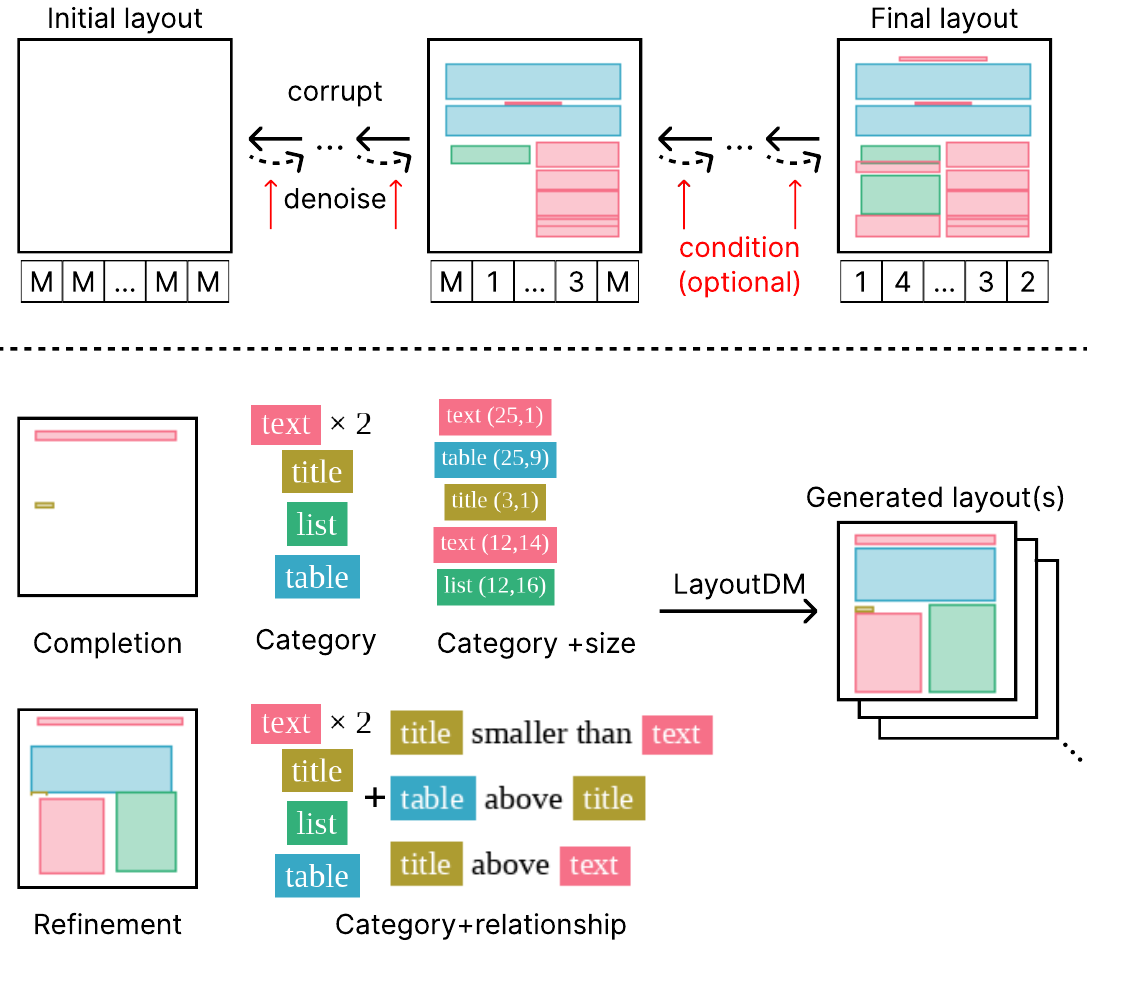}
    \caption{
        Overview of LayoutDM.
        \textbf{Top}: LayoutDM is trained to gradually generate a complete layout from a blank state in discrete state space.
        \textbf{Bottom}: During sampling, we can steer LayoutDM to perform various conditional generation tasks without additional training or external models.
    }
    \label{fig:teaser}
\end{figure}

In this paper, we propose to utilize discrete state-space diffusion models~\cite{hoogeboom2021argmax,austin2021structured,gu2022vector} for layout generation tasks.
Diffusion models have shown promising performance for various generation tasks, including images and texts~\cite{ho2020denoising}.
We formulate the diffusion process for layout structure by \emph{modality-wise discrete diffusion}, and train a denoising backbone network to progressively infer the complete layout with or without conditional inputs.
To support variable-length layout data, we extend the discrete state-space with a special \texttt{PAD} token instead of the typical end-of-sequence token used in autoregressive models.
Our model can incorporate complex layout constraints via \textit{logit adjustment}, so that we can refine an existing layout or impose relative size constraints between elements without additional training.

We discuss two key advantages of LayoutDM over existing models for conditional layout generation.
Our model avoids the immutable dependency chain issue~\cite{kong2022blt} that happens in autoregressive models~\cite{gupta2021layout}.
Autoregressive models fail to perform conditional generation when the condition disagrees with the pre-defined generation order of elements and attributes.
Unlike non-autoregressive models~\cite{kong2022blt}, our model can generate variable-length elements.
We empirically show in \cref{sec:quantitative_evaluation} that naively extending non-autoregressive models by padding results in suboptimal variable length generation while padding combined with our diffusion formulation leads to significant improvement.

We evaluate LayoutDM on various layout generation tasks tackled by previous works~\cite{kong2022blt,lee2020neural,rahman2021ruite,paschalidou2021atiss} using two large-scale datasets, Rico~\cite{deka2017rico} and PubLayNet~\cite{zhong2019publaynet}.
LayoutDM outperforms task-agnostic baselines in the majority of cases and shows promising performance compared with task-specific baselines.
We further conduct an ablation study to prove the significant impact of our design choices in LayoutDM, including quantization of continuous variables and positional embedding.

We summarize our contributions as follows:
\begin{itemize}[noitemsep,nolistsep,leftmargin=*]
    \item We formulate the discrete diffusion process for layout generation and propose a modality-wise diffusion and a padding approach to model highly structured layout data.
    \item We propose to inject complex layout constraints via masking and logit adjustment during the inference,
    so that our model can solve diverse tasks in a single model.
    \item We empirically show solid performance for various conditional layout generation tasks on public datasets.
\end{itemize}

\section{Related Work}
\subsection{Layout Generation}
Studies on automatic layout generation have appeared several times in literature~\cite{lok2001survey,agrawala2011design,o2014learning,yang2016automatic}.
Layout tasks are commonly observed in design applications, including magazine covers, posters, presentation slides, application user interface, or banner advertising~\cite{deka2017rico, yang2016automatic,zheng2019content,qian2020retrieve,yamaguchi2021canvasvae,guo2021vinci,kikuchi2021modeling,fu2022doc2ppt}.
Recent approaches to layout generation consider both unconditional generation~\cite{jyothi2019layoutvae,gupta2021layout,arroyo2021variational,jiang2022coarse} and conditional generation in various setups, such as conditional inputs of category or size~\cite{li2019layoutgan,lee2020neural,Kikuchi2021,kong2022blt}, relational constraints~\cite{lee2020neural,Kikuchi2021}, element completion~\cite{gupta2021layout}, and refinement~\cite{rahman2021ruite}.
Some attempt at solving multiple tasks in a single model~\cite{kong2022blt,paschalidou2021atiss}.

BLT~\cite{kong2022blt} points out that the recent autoregressive decoders~\cite{arroyo2021variational,gupta2021layout} are not fully capable of considering partial inputs, \ie known elements or attributes, during generation because they have a fixed generation order.
BLT addresses the conditional generation by fill-in-the-blank task formulation using a bidirectional Transformer encoder similar to masked language models~\cite{devlin-etal-2019-bert}.
However, BLT cannot solve layout completion demonstrated in the decoder-based models because of the requirement of the known number of elements.
Our LayoutDM enjoys the best of both worlds and supports a broader range of conditional generation tasks in a single model.

Another layout-specific consideration is the complex user-specified constraints, such as the positional requirements between two boxes (e.g., a header box should be on top of a paragraph box).
Earlier approaches~\cite{merrell2011interactive,yu2011make,o2014learning} propose hand-crafted cost functions representing the violation degree of aesthetic constraints so that those constraints guide the optimization process of layout inference.
CLG-LO~\cite{Kikuchi2021} proposes an aesthetically constrained optimization framework for pre-trained GANs.
Our LayoutDM solves such constrained generation tasks on top of the task-agnostic iterative prediction via logit adjustment.

\subsection{Discrete Diffusion Models}
Diffusion models~\cite{sohl2015deep} are generative models characterized by a forward and reverse Markov process.
The forward process corrupts the data into a sequence of increasingly noisy variables.
The reverse process gradually denoises the variables toward the actual data distribution.
Diffusion models are stable to train and achieve faster sampling than autoregressive models by parallel iterative refinement.
Recently, many approaches have learned the reverse process by a neural network and show strong empirical performance~\cite{ho2020denoising,song2020improved,dhariwal2021diffusion} in continuous state spaces, such as images.

Discrete state spaces are a natural representation of discrete variables, such as text. D3PM~\cite{austin2021structured} extends the pioneering work of Hoogeboom~\etal~\cite{hoogeboom2021argmax} to structured categorical corruption processes for diffusion models in discrete state spaces, while maintaining the advantages of diffusion models for continuous state spaces.
VQDiffusion~\cite{gu2022vector} develops a corruption approach called mask-and-replace, so as to avoid accumulated prediction errors that are common in models based on iterative prediction.
Following the corruption model of VQDiffusion, we carefully design a modality-wise corruption process for layout tasks that involve tokens from disjoint sets of vocabulary per modality.

Several studies consider a conditional input to the inference process of diffusion models.
Some approaches alter the reverse diffusion iteration to carefully inject given conditions for free-form image inpainting~\cite{lugmayr2022repaint} or image editing by strokes or composition~\cite{meng2022sdedit}.
We extend the discrete state-space diffusion models via hard masking or logit adjustment to support the conditional generation of layouts.

\section{LayoutDM}
Our LayoutDM builds on discrete-state space diffusion models~\cite{austin2021structured,gu2022vector}.
We first briefly review the fundamental of discrete diffusion models in \cref{subsec:discrete_diffusion}.
\cref{subsec:layout_diffusion_unconditional} explains our approach to layout generation within the diffusion framework while discussing features inherent in layout compared with text.
\cref{subsec:layout_diffusion_conditional} discusses how we extend denoising steps to perform various conditional layout generation by imposing conditions in each step of the reverse process.

\subsection{Preliminary: Discrete Diffusion Models}
\label{subsec:discrete_diffusion}

Diffusion models~\cite{sohl2015deep} are generative models characterized by a forward and reverse Markov process.
While many diffusion models are defined on continuous space with Gaussian corruption, D3PM~\cite{austin2021structured} introduces a general diffusion framework for categorical variables designed primarily for texts.
Let $T \in \mathbb{N}$ be a total timestep of the diffusion model, we first explain the forward diffusion process.
For a scalar discrete variable with $K$ categories $z_{t} \in \{1,2,\ldots,K\}$ at timestep $t \in \mathbb{N}$, probabilities that $z_{t-1}$ transits to $z_{t}$ are defined by using a transition matrix $\bm{Q}_{t} \in [0,1]^{K \times K}$, with $[Q_{t}]_{mn} = q(z_{t}\!=\!m | z_{t-1}\!=\!n)$,

\begin{equation}
q(z_{t}|z_{t-1}) = \bm{v}(z_{t})^{\!\top} \mathbf{Q}_{t} \bm{v}(z_{t-1}),
\end{equation}
where $\bm{v}(z_{t}) \in \{0,1\}^{K}$ is a column one-hot vector of $z_{t}$.
The categorical distribution over $z_{t}$ given $z_{t-1}$ is computed by a column vector $\mathbf{Q}_{t} \bm{v}(z_{t-1}) \in [0,1]^{K}$.
Assuming the Markov property,
we can derive $q(z_{t}|z_{0}) = \bm{v}(z_{t})^{\!\top}\overline{\mathbf{Q}}_{t}\bm{v}(z_{0})$ where $\overline{\mathbf{Q}}_{t}\!=\!\mathbf{Q}_{t}\mathbf{Q}_{t-1}\cdots\mathbf{Q}_{1}$ and:
\begin{align}
    &q(z_{t-1}|z_{t}, z_{0}) = \frac{
        q(z_{t}|z_{t-1}, z_{0})\,q(z_{t-1}|z_{0})
    }{
        q(z_{t}|z_{0})
    } \nonumber \\
    &= \frac{
        \left(\bm{v}\!\left(z_{t}\right)^{\!\top}\!\mathbf{Q}_{t}\bm{v}\!\left(z_{t-1}\right)\right)
        \left( \bm{v}\!\left(z_{t-1}\right)^{\!\top}\overline{\mathbf{Q}}_{t-1}\bm{v}\!\left(z_{0}\right) \right)
    }{
        \bm{v}\!\left(z_{t}\right)^{\!\top}\overline{\mathbf{Q}}_{t}\bm{v}\!\left(z_{0}\right)
    }. \label{eq:q_posterior}
\end{align}
Note that due to the Markov property, $q(z_{t}|z_{t-1}, z_{0})=q(z_{t}|z_{t-1})$.
When we consider $N$-dimensional variables $\bm{z}_{t} \in \{1,2,\ldots,K\}^{N}$, the corruption is applied to each variable $z_{t}$ independently.
In the following, we explain with $N$-dimensional variables $\bm{z}_{t}$.

In contrast to the forward process, the reverse denoising process considers a conditional distribution of $\bm{z}_{t-1}$ over $\bm{z}_{t}$ by a neural network $p_{\theta}(\bm{z}_{t-1}|\bm{z}_{t}) \in [0,1]^{N \times K}$.
$\bm{z}_{t-1}$ is sampled according to this distribution.
Note that the typical implementation is to predict unnormalized log probabilities $\log p_{\theta}(\bm{z}_{t-1}|\bm{z}_{t})$ by a stack of bidirectional Transformer encoder blocks.
D3PM uses a neural network $\tilde{p}_{\theta}(\tilde{\bm{z}}_{0}|\bm{z}_{t})$, combines it with the posterior $q(\bm{z}_{t-1}|\bm{z}_{t},\bm{z}_{0})$, and sums over possible $\tilde{\bm{z}_{0}}$ to obtain the following parameterization:
\begin{equation}
p_{\theta}(\bm{z}_{t-1}|\bm{z}_{t}) \propto \sum_{\tilde{\bm{z}}_{0}}q(\bm{z}_{t-1}|\bm{z}_{t},\tilde{\bm{z}}_{0})~\tilde{p}_{\theta}(\tilde{\bm{z}}_{0}|\bm{z}_{t}). \label{eq:single_step_in_inference}
\end{equation}

In addition to the commonly used variational lower bound objective $\mathcal{L}_\mathrm{vb}$, D3PM introduces an auxiliary denoising objective. The overall objective is as follows:
\begin{equation}
    \mathcal{L}_{\lambda} = \mathcal{L}_\mathrm{vb} + \lambda
    \underset{\substack{ \bm{z}_{t} \sim q(\bm{z}_{t}|\bm{z}_{0}) \\ \bm{z}_{0} \sim q(\bm{z}_{0})}}{\mathbb{E}}
    \left[ -\log \tilde{p}_{\theta}\left(\bm{z}_{0}|\bm{z}_{t}\right) \right], \label{eq:total_loss}
\end{equation}
where $\lambda$ is a hyper-parameter to balance the two loss terms.

Although D3PM proposes many variants of $\mathbf{Q}_{t}$, VQDiffusion~\cite{gu2022vector} offers an improved version of $\mathbf{Q}_{t}$ called mask-and-replace strategy.
They introduce an additional special token \texttt{[MASK]}
and three probabilities $\gamma_{t}$ of replacing the current token with the \texttt{[MASK]} token, $\beta_{t}$ of replacing the token with other tokens, and $\alpha_{t}$ of not changing the token.
The \texttt{[MASK]} token never transitions to other states.
The transition matrix $\mathbf{Q}_{t} \in [0,1]^{(K+1)\times(K+1)}$ is defined by:
\begin{equation}
\mathbf{Q}_{t} = \begin{bmatrix}
\alpha_{t}+\beta_{t} & \beta_{t} & \cdots & \beta_{t} & 0  \\
\beta_{t} & \alpha_{t}+\beta_{t} & \cdots & \beta_{t} & 0  \\
\vdots & \vdots & \ddots & \beta_{t} & 0  \\
\beta_{t} & \beta_{t} & \beta_{t} & \alpha_{t}+\beta_{t} &  0  \\
\gamma_{t} & \gamma_{t} & \gamma_{t} & \gamma_{t} & 1 \\
\end{bmatrix}.
\label{eq:Q_mask_and_replace}
\end{equation}
$(\alpha_{t}, \beta_{t}, \gamma_{t})$ is carefully designed so that $z_{t}$ converges to the \texttt{[MASK]} token for sufficiently large $t$.
During testing, we start from $\bm{z}_{T}$ filled with \texttt{[MASK]} tokens and iteratively sample new set of tokens $\bm{z}_{t-1}$ from $p_{\theta}(\bm{z}_{t-1}|\bm{z}_{t})$.

\subsection{Unconditional Layout Generation}
\label{subsec:layout_diffusion_unconditional}

\begin{figure}[t]
    \centering
    \includegraphics[width=\hsize]{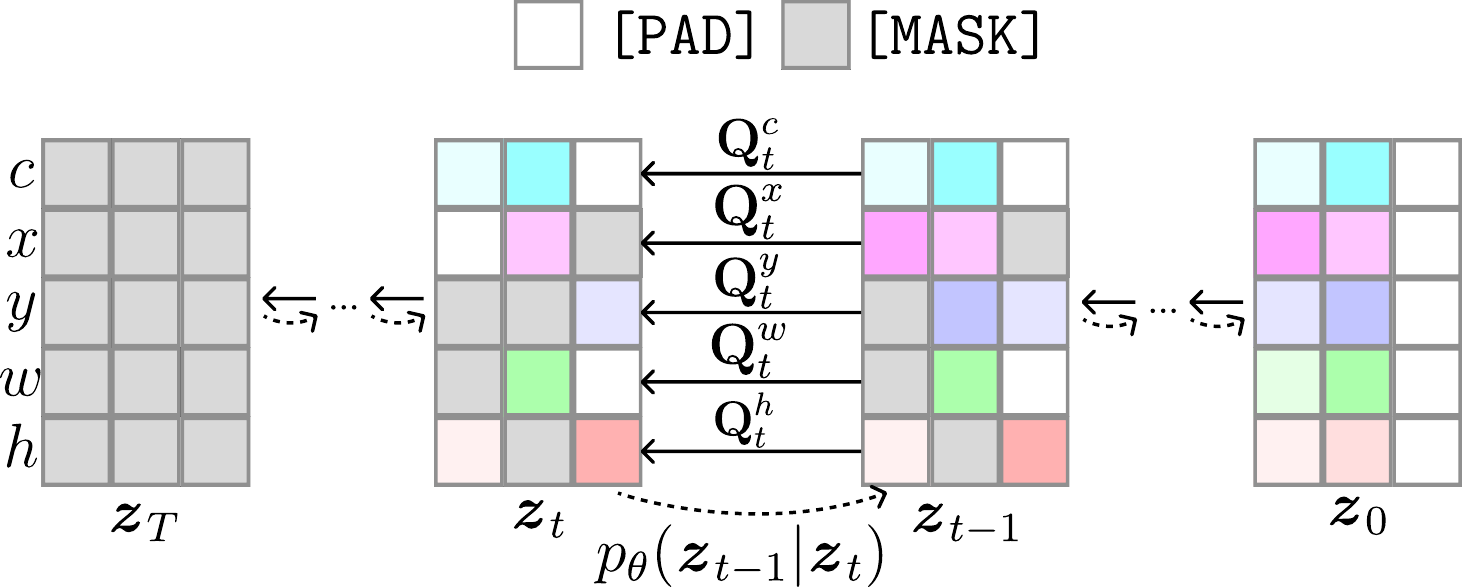}
    \caption{
        Overview of the corruption and denoising processes in LayoutDM.
        For simplicity, we use a toy layout consisting of two elements and the model generates three elements at maximum.
    }
    \label{fig:overview}
\end{figure}

A layout $l$ is a set of elements represented by $l = \left\{\left(c_{1}, \bm{b}_{1}\right), \ldots, \left(c_{E}, \bm{b}_{E}\right) \right\}$. $E \in \mathbb{N}$ is the number of elements in the layout. $c_{i} \in \{1, \ldots, C\}$ is categorical information of the $i$-th element in the layout. $\bm{b}_{i} \in [0,1]^4$ is the bounding box of the $i$-th element in normalized coordinates, where the first two values indicate the center location, and the last two indicate the width and height.
Following previous works~\cite{arroyo2021variational,gupta2021layout,kong2022blt} that regard layout generation as generating a sequence of tokens, we
quantize each value in $\bm{b}_{i}$ and obtain $[x_{i}, y_{i}, w_{i}, h_{i}]^\top \in \{1, \ldots, B\}^4$, where $B$ is a number of the bins. The layout $l$ is now represented by $l = \left\{\left(c_{1}, x_{1}, y_{1}, w_{1}, h_{1}\right), \ldots \right\}$.

In this work, we corrupt a layout in a modality-wise manner in the forward process, and we denoise the corrupted layout while considering all elements and modalities in the reverse process, as we illustrate in \cref{fig:overview}.
Similarly to D3PM~\cite{austin2021structured}, we parameterize $p_{\theta}$ by a Transformer encoder~\cite{vaswani2017attention}, which processes an ordered 1D sequence.
To process $l$ by $p_{\theta}$ while avoiding the order dependency issue~\cite{kong2022blt}, we randomly shuffle $l$ in element-wise manner and then flatten it to produce $l_\mathrm{flat} = (c_{1}, x_{1}, y_{1}, w_{1}, h_{1}, c_{2}, x_{2}, \ldots )$.

\paragraph{Variable length generation}
Existing diffusion models generate fixed-dimensional data and are not directly applicable to the layout generation because the number of elements in each layout varies.
To handle this, we introduce a \texttt{[PAD]} token and define a maximum number of elements in the layout as $M \in \mathbb{N}$.
Each layout is fixed-dimensional data composed of $5M$ tokens by appending $5(M-E)$ \texttt{[PAD]} tokens.
\texttt{[PAD]} is treated similarly to the ordinary token in VQDiffusion and $\mathbf{Q}_{t}$'s dimension becomes $(K+2) \times (K+2)$.

\paragraph{Modality-wise diffusion}
Discrete state-space models assume that all the standard tokens are switchable by corruption. However, layout tokens comprise a disjoint set of token groups for each attribute in the element. For example, applying the transition rule \cref{eq:Q_mask_and_replace} may change a token representing an element's category to another token representing the width.
To avoid such invalid switching, we propose to apply disjoint corruption matrices $\mathbf{Q}_{t}^{c},\mathbf{Q}_{t}^{x},\mathbf{Q}_{t}^{y},\mathbf{Q}_{t}^{w},\mathbf{Q}_{t}^{h}$ for tokens representing different attributes $c, x, y, w, h$, as we show in \cref{fig:overview}. The size of each matrix is $(C+2) \times (C+2)$ for $\mathbf{Q}_{t}^{c}$ and otherwise $(B+2) \times (B+2)$, where $+2$ is for \texttt{[{PAD}]} and \texttt{[{MASK}]}.

\paragraph{Adaptive Quantization}
The distribution of the position and size information in layouts is highly imbalanced; e.g., elements tend to be aligned to either left, center, or right.
Applying uniform quantization to those quantities as in existing layout generation models~\cite{arroyo2021variational,gupta2021layout,kong2022blt} results in the loss of information.
As a pre-processing, we propose to apply a classical clustering algorithm, such as KMeans~\cite{macqueen1967classification} on $x$, $y$, $w$, and $h$ independently to obtain balanced position and size tokens for each dataset.
We show in \cref{sec:ablation_study} how quantization strategy affects the resulting quality.

\paragraph{Decoupled Positional Encoding}
Previous works apply standard positional encoding to a flattened sequence of layout tokens $l_\mathrm{flat}$~\cite{arroyo2021variational,gupta2021layout,kong2022blt}.
We argue that this flattening approach could lose the structure information of the layout and lead to inferior generation performance.
In layout, each token has two types of indices: $i$-th element and $j$-th attribute.
We empirically find that independently applying positional encoding to those indices improves final generation performance, which we study in \cref{sec:ablation_study}.

\subsection{Conditional Generation}
\label{subsec:layout_diffusion_conditional}
We elaborate on solving various conditional layout generation tasks using pre-trained frozen LayoutDM.
We inject conditional information in both the initial state $\bm{z}_{T}$ and sampled states $\{\bm{z}_{t}\}_{t=0}^{T-1}$ during inference but do not modify the denoising network $p_{\theta}$.
The actual implementation of the injection differs by the type of conditions.

\paragraph{Strong Constraints}
The most typical condition is partially known layout fields.
Let $\bm{z}^\mathrm{known} \in \mathbb{Z}^{N}$ contain the known fields and $\bm{m} \in \{0, 1\}^{N}$ be a mask vector denoting the known and unknown field as $1$ and $0$, respectively. In each timestep $t$, we sample $\hat{\bm{z}}_{t-1}$ from $p_{\theta}(\bm{z}_{t-1}|\bm{z}_{t})$ in \cref{eq:single_step_in_inference} and then inject the condition by $\bm{z}_{t-1} = \bm{m} \odot \bm{z}^\mathrm{known} + (\bm{1} - \bm{m}) \odot \hat{\bm{z}}_{t-1}$, where $\bm{1}$ denotes a $N$-dimensional all-ones vector and $\odot$ denotes element-wise product.

\paragraph{Weak Constraints}
We may impose a weaker constraint during generation, such as an element in the center. We offer a way to impose such constraints in a unified framework without additional training or external neural network models.
We propose to adjust the logits to inject weak constraints in log probability space by
\begin{equation}
    \log \hat{p}_{\theta}(\bm{z}_{t-1}|\bm{z}_{t}) \propto \log p_{\theta}(\bm{z}_{t-1}|\bm{z}_{t}) + \lambda_{\pi} \bm{\pi}, \\ \label{eq:prior_addition}
\end{equation}
where $\bm{\pi} \in \mathbb{R}^{N \times K}$ is a prior term that weights the desired outputs, and $\lambda_{\pi} \in \mathbb{R}$ is a hyper-parameter.
The prior term can be defined either hard-coded (Refinement in \cref{sec:quantitative_evaluation}) or through differentiable loss functions (Relationship in \cref{sec:quantitative_evaluation}).
Let $\{\mathcal{L}_i\}_{i=1}^L$ be a set of differentiable loss functions given the prediction, the later prior definition can be written by:
\begin{equation}
    \bm{\pi} = -\nabla_{p_{\theta}(\bm{z}_{t-1}|\bm{z}_{t})} \sum_{i=1}^{L} \mathcal{L}_{i}\left(p_{\theta}\left(\bm{z}_{t-1}|\bm{z}_{t}\right)\right). \\ \label{eq:prior_addition_by_gradient}
\end{equation}
Although the formulation of \cref{eq:prior_addition_by_gradient} resembles steering diffusion models by gradients from external models ~\cite{dhariwal2021diffusion,liu2022compositional}, our primal focus is incorporating classical hand-crafted energies for aesthetics principles of layout~\cite{o2014learning} that do not depend on an external model.
In practice, we tune the hyper-parameters for imposing weak constraints, such as $\lambda_{\pi}$.
Note that these hyper-parameters are only for inference and are easier to tune than the other training hyper-parameters.

\section{Experiment}
\subsection{Datasets}
We use two large-scale datasets for comparison, Rico~\cite{deka2017rico} and PubLayNet~\cite{zhong2019publaynet}.
As we mention in \cref{subsec:layout_diffusion_unconditional}, an element in a layout for each dataset is described by the five attributes.
For preprocessing, we set the maximum number of elements per layout $M$ to 25. If a layout contains more elements, we discard the whole layout.

We provide an overview of each dataset.
\textbf{Rico} is a dataset of user interface designs for mobile applications containing 25 element categories such as text button, toolbar, and icon.
We divide the dataset into 35,851 / 2,109 / 4,218 samples for train, validation, and test splits.
\textbf{PubLayNet} is a dataset of research papers containing five element categories, such as table, image, and text. We divide the dataset into 315,757 / 16,619 / 11,142 samples for train, validation, and test splits.

\subsection{Evaluation Metrics}
We employ two primary metrics: FID and Maximum IoU (Max.).
These metrics take into account both fidelity and diversity~\cite{heusel2017gans}, which are two mutually complementary properties widely used in evaluating generative models.
FID~\cite{heusel2017gans} captures the similarity of generated data to real ones in feature space. We employ an improved feature extraction model for layouts \cite{Kikuchi2021} instead of a conventional method~\cite{lee2020neural} to compute FID.
Maximum IoU~\cite{Kikuchi2021} measures the conditional similarity between generated and real layouts.
The similarity is measured by computing optimal matching that maximizes average IoU between generated and real layouts that have an identical set of categories.
For reference, we show the FID and Maximum IoU computed between the validation and test data as~\emph{Real data}.

\subsection{Tasks and Baselines}
We test LayoutDM on six tasks for evaluation.

\noindent
\textbf{Unconditional} generates layouts without any conditional input or constraint. \\
\textbf{Category$\rightarrow$size+position (C$\rightarrow$S+P)} is a generation task conditioned on the category of each element~\cite{kong2022blt}. \\
\textbf{Category+size$\rightarrow$ position (C+S$\rightarrow$P)} is conditioned on the category and size of each element. \\
\textbf{Completion} is conditioned on a small number of elements whose attributes are all known. Given a complete layout, we randomly sample from $0\%$ to $20\%$ of elements. \\
\textbf{Refinement} is conditioned on a noisy layout in which only geometric information is perturbed~\cite{rahman2021ruite}.
Following RUITE~\cite{rahman2021ruite}, we synthesize the input layout by adding random noise to the size and position of each element.
We sample noise from a standard normal distribution with a mean of 0 and a variance of 0.01. \\
\textbf{Relationship} is conditioned on the category of each element and some relationship constraints between the elements~\cite{lee2020neural}. Following CLG-LO~\cite{Kikuchi2021}, we employ the size and location relationships and randomly sample $10\%$ relationships between elements for the experiment.

The first four tasks handle basic layout fields.
We include a few task-agnostic models for comparison using existing controllable layout generation methods or simple adaptation of generative models in the following:

\noindent \textbf{LayoutTrans} is a simple autoregressive model~\cite{gupta2021layout} trained on a element-level shuffled layout, following \cite{paschalidou2021atiss}.
We set a variable generation order to c$\rightarrow$w$\rightarrow$h$\rightarrow$x$\rightarrow$y. \\
\noindent \textbf{MaskGIT$^\ast$} is originally a non-autoregressive model for unconditional fixed-length data generation~\cite{chang2022maskgit}.
We use \texttt{[PAD]} to enable variable-length generation. \\
\noindent \textbf{BLT} is a non-autoregressive model with layout-specific decoding strategy~\cite{kong2022blt}. \\
\noindent \textbf{BART} is a denoising autoencoder that can solve both comprehension and generation tasks based on Transformer encoder-decoder backbone~\cite{lewis2020bart}.
We randomly generate a number of \texttt{[MASK]} tokens from a uniform distribution between one and the sequence length, and perform masking based on the number. \\
\noindent \textbf{VQDiffusion$^\ast$} is a diffusion-based model originally for text-to-image generation~\cite{gu2022vector}.
We adapt the model for layout using $K=C+4B+2$ tokens, including \texttt{[PAD]}.

\subsection{Implementation Details}
We re-implement most of the models since there are few official implementations publicly available except~\cite{gupta2021layout,Kikuchi2021,kong2022blt}\footnote{Unfortunately, most datasets have no official train-val-test splits, and previous approaches work on different splits and pre-processing strategies. Furthermore, models for FID computation also vary. Thus, we cannot directly compare our results with the reported figure in the literature.}.
We train all the models on the two datasets with three independent trials and report the average of the results.

LayoutDM follows VQDiffusion for hyper-parameters unless specified, such as configurations for $p_{\theta}$ and the transition matrix parameters \ie $\alpha_{t}$ and $\gamma_{t}$.
We set the loss weight $\lambda=0.1$ (in \cref{eq:total_loss}) and the diffusion timesteps $T=100$.
For optimization, we use AdamW~\cite{loshchilov2019decoupled} with learning rate of $5.0 \times 10^{-4}$, $\beta_{1}=0.9$, and $\beta_{2}=0.98$.

Many models, including LayoutDM, use Transformer~\cite{vaswani2017attention} encoder backbone.
We define a shared configuration as follows:
4 layers, 8 attention heads, 512 embedding dimensions, 2048 hidden dimensions, and $0.1$ dropout rate.
For other models with extra modules, we adjust the number of hidden dimensions to roughly match the number of parameters for a fair comparison.
We randomly shuffle elements in the layout to avoid fixed-order generation during training.
We search best hyper-parameters to obtain the best FID using the validation set.

{
\setlength{\tabcolsep}{3.5pt}
\begin{table*}
\centering
\caption{
    Quantitative comparison in conditional generation given partially known fields.
    Top two results are highlighted in \textbf{bold} and \underline{underline}, respectively.
    $^{\dagger}$ indicates the results of BLT trained with \texttt{[PAD]} as an additional vocabulary since the original model cannot perform unordered completion in practice.
}
\label{tab:comparison_inpainting}
\vspace*{-2mm}

\begin{tabular}{lcccccccccccc} \toprule
\multicolumn{1}{r}{Task} & \multicolumn{4}{c}{Category $\rightarrow$ Size+Position} & \multicolumn{4}{c}{Category+Size $\rightarrow$ Position} & \multicolumn{4}{c}{Completion} \\
\cmidrule(lr){2-5} \cmidrule(lr){6-9} \cmidrule(lr){10-13}
\multicolumn{1}{r}{Dataset} & \multicolumn{2}{c}{Rico} & \multicolumn{2}{c}{PubLayNet} & \multicolumn{2}{c}{Rico} & \multicolumn{2}{c}{PubLayNet} & \multicolumn{2}{c}{Rico} & \multicolumn{2}{c}{PubLayNet} \\
Model
& FID $\downarrow$ & Max. $\uparrow$ & FID $\downarrow$ & Max. $\uparrow$
& FID $\downarrow$ & Max. $\uparrow$ & FID $\downarrow$ & Max. $\uparrow$
& FID $\downarrow$ & Max. $\uparrow$ & FID $\downarrow$ & Max. $\uparrow$ \\
\midrule

\hspace{1mm}{\scriptsize Task-specific models} \\

LayoutVAE~\cite{jyothi2019layoutvae}
& 33.3\std{4.0} & 0.249\std{1.1} & 26.0\std{2.1} & 0.316\std{2.2}
& 30.6\std{2.8} & 0.283\std{0.1} & 27.5\std{4.2} & 0.315\std{0.7}
& - & - & - & - \\

NDN-none~\cite{lee2020neural}
& 28.4\std{8.7} & 0.158\std{4.5} & 61.1\std{15.9} & 0.162\std{6.8}
& 62.8\std{7.3} & 0.219\std{1.2} & 69.4\std{18.1} & 0.222\std{3.0}
& - & - & - & - \\

LayoutGAN++~\cite{Kikuchi2021}
& 6.84\std{15.8} & \underline{0.267}\std{1.1} & 24.0\std{51.4} & 0.263\std{11.1}
& 6.22\std{20.1} & \underline{0.348}\std{1.2} & 9.94\std{5.3} & 0.342\std{1.4}
& - & - & - & -
\\

\hspace{1mm}{\scriptsize Task-agnostic models} \\

LayoutTrans~\cite{gupta2021layout} %
& 5.57\std{11.0} & 0.223\std{1.0} & 14.1\std{1.4} & 0.272\std{2.1}
& 3.73\std{4.1} & 0.323\std{1.6} & 16.9\std{4.7} & 0.320\std{0.9}
& \textbf{3.71}\std{5.7} & 0.537\std{0.3} & \underline{8.36}\std{2.3} & \underline{0.451}\std{0.1}
\\

MaskGIT$^\ast$~\cite{chang2022maskgit}
& 26.1\std{5.6} & 0.262\std{1.9} & 17.2\std{5.4} & \underline{0.319}\std{2.7}
& 8.05\std{17.0} & 0.320\std{1.2} & 5.86\std{6.0} & 0.380\std{0.5}
& 33.5\std{4.0} & 0.533\std{2.9} & 19.7\std{2.5} & \textbf{0.484}\std{0.7}\\

BLT~\cite{kong2022blt}
& 17.4\std{6.7} & 0.202\std{3.1} & 72.1\std{14.7} & 0.215\std{7.6}
& 4.48\std{4.2} & 0.340\std{0.8} & \underline{5.10}\std{1.1} & \textbf{0.387}\std{0.4}
& 117\std{6.1}$^{\dagger}$ & 0.471\std{25.7}$^{\dagger}$ & 131\std{17.4}$^{\dagger}$ & 0.345\std{1.9}$^{\dagger}$ \\

BART~\cite{lewis2020bart} %
& \underline{3.97}\std{6.2} & 0.253\std{1.1} & \underline{9.36}\std{1.5} & \textbf{0.320}\std{0.5}
& \underline{3.18}\std{2.2} & 0.334\std{0.3} & 5.88\std{3.5} & 0.375\std{0.3}
& \underline{8.87}\std{26.4} & 0.527\std{8.1} & 9.58\std{5.8} & 0.446\std{1.3} \\

VQDiffusion$^\ast$~\cite{gu2022vector}
& 4.34\std{5.5} & 0.252\std{1.3} & 10.3\std{4.2} & \underline{0.319}\std{1.8}\
& 3.21\std{2.9} & 0.331\std{1.2} & 7.13\std{3.4} & 0.374\std{0.2}
& 11.0\std{11.8} & \underline{0.541}\std{3.3} & 11.1\std{5.3} & 0.373\std{0.9} \\

LayoutDM
& \textbf{3.55}\std{5.0} & \textbf{0.277}\std{0.3} & \textbf{7.95}\std{1.7} & 0.310\std{0.3}
& \textbf{2.22}\std{1.6} & \textbf{0.392}\std{0.1} & \textbf{4.25}\std{1.2} & \underline{0.381}\std{0.2}
& 9.00\std{1.9} & \textbf{0.576}\std{2.9} & \textbf{7.65}\std{1.4} & 0.377\std{0.6} \\

\midrule

Real data & 1.85 & 0.691 & 6.25 & 0.438 & 1.85 & 0.691 & 6.25 & 0.438 & 1.85 & 0.691 & 6.25 & 0.438 \\

\bottomrule \end{tabular}
\end{table*}
\setlength\tabcolsep{6pt}
}
{
    \setlength\tabcolsep{1pt}
    \begin{figure*}[t]
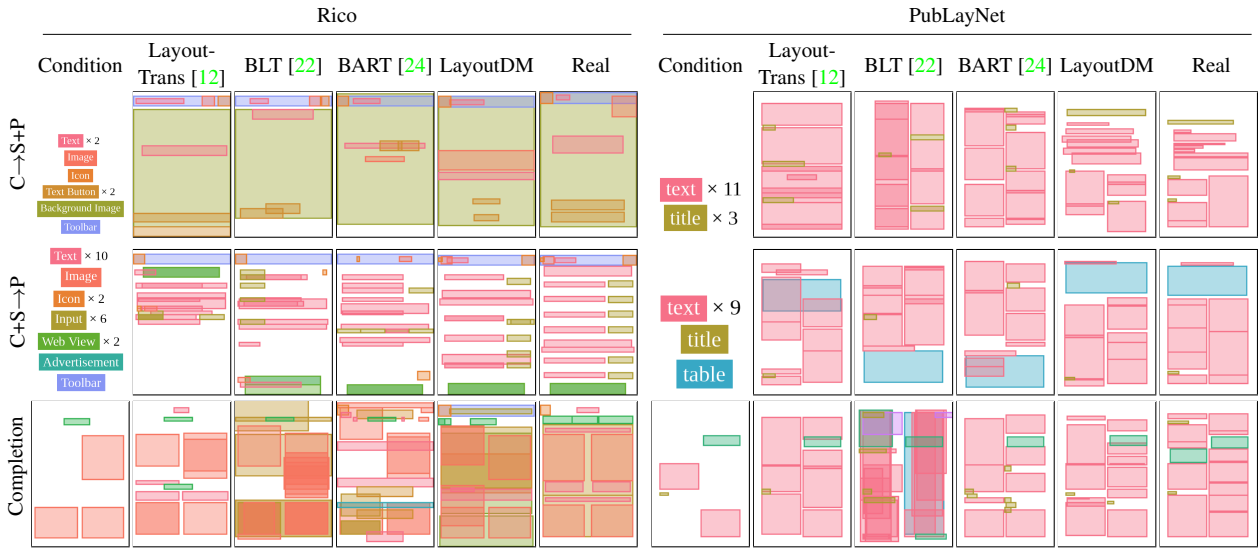

    \centering
    \newcommand{\length}{0.11}
    \newcommand{\figcmd}[3]{
        \includegraphics[width=0.07\hsize]{images/comparison/#1/#2/#3_input.png} &
        \frame{\includegraphics[height=\length\hsize]{images/comparison/#1/#2/#3_layouttrans.png}} &
        \frame{\includegraphics[height=\length\hsize]{images/comparison/#1/#2/#3_blt.png}} &
        \frame{\includegraphics[height=\length\hsize]{images/comparison/#1/#2/#3_bart.png}} &
        \frame{\includegraphics[height=\length\hsize]{images/comparison/#1/#2/#3_layoutdm.png}} &
        \frame{\includegraphics[height=\length\hsize]{images/comparison/#1/#2/#3_gt.png}}
    }
    \newcommand{\figcmdtwo}[3]{
        \frame{\includegraphics[height=\length\hsize]{images/comparison/#1/#2/#3_input.png}} &
        \frame{\includegraphics[height=\length\hsize]{images/comparison/#1/#2/#3_layouttrans.png}} &
        \frame{\includegraphics[height=\length\hsize]{images/comparison/#1/#2/#3_blt.png}} &
        \frame{\includegraphics[height=\length\hsize]{images/comparison/#1/#2/#3_bart.png}} &
        \frame{\includegraphics[height=\length\hsize]{images/comparison/#1/#2/#3_layoutdm.png}} &
        \frame{\includegraphics[height=\length\hsize]{images/comparison/#1/#2/#3_gt.png}}
    }
    \footnotesize{
        \begin{tabular}{cccccccccccccc}
            & \multicolumn{6}{c}{Rico} & & \multicolumn{6}{c}{PubLayNet} \\
            \cmidrule(lr){2-7} \cmidrule(lr){9-14}
            & \multirow{2}{*}{Condition} & \multirow{2}{*}{\shortstack{Layout-\\Trans~\cite{gupta2021layout}}} & \multirow{2}{*}{BLT~\cite{kong2022blt}} & \multirow{2}{*}{BART~\cite{lewis2020bart}} & \multirow{2}{*}{\shortstack{LayoutDM}} & \multirow{2}{*}{Real} & & \multirow{2}{*}{Condition} & \multirow{2}{*}{\shortstack{Layout-\\Trans~\cite{gupta2021layout}}} & \multirow{2}{*}{BLT~\cite{kong2022blt}} & \multirow{2}{*}{BART~\cite{lewis2020bart}} & \multirow{2}{*}{\shortstack{LayoutDM}} & \multirow{2}{*}{Real} \\
            & & & & & & & & & & \\
            \raisebox{3.0\normalbaselineskip}[0pt][0pt]{\rotatebox[origin=c]{90}{C$\rightarrow$S+P}} & \figcmd{rico25}{c}{45869} & \hfill & \figcmd{publaynet}{c}{PMC3335522_00000} \\
            \raisebox{3.0\normalbaselineskip}[0pt][0pt]{\rotatebox[origin=c]{90}{C+S$\rightarrow$P}} & \figcmd{rico25}{cwh}{6539} & \hfill & \figcmd{publaynet}{cwh}{PMC5153480_00002} \\
            \raisebox{3.0\normalbaselineskip}[0pt][0pt]{\rotatebox[origin=c]{90}{Completion}} & \figcmdtwo{rico25}{partial}{25440} & \hfill & \figcmdtwo{publaynet}{partial}{PMC4055381_00001} \\
        \end{tabular}
    }
    \caption{Comparison in conditional generation given partially known fields.}
    \label{fig:comparison_inpainting}
    \end{figure*}
}

{
\setlength{\tabcolsep}{3.5pt}
\begin{table}
    \centering

    \caption{
        Quantitative comparison in unconditional generation.
        Top two results are highlighted in \textbf{bold} and \underline{underline}, respectively.
    }
    \label{tab:comparison_unconditional}
    \vspace*{-2mm}
    \newcommand{\cg}[1]{#1}

    \begin{tabular}{lcccc} \toprule
        \multicolumn{1}{r}{Dataset} & \multicolumn{2}{c}{Rico} & \multicolumn{2}{c}{PubLayNet} \\
        Model & FID $\downarrow$ & Align. $\downarrow$ & FID $\downarrow$ & Align. $\downarrow$ \\ \midrule

        \cg{LayoutTrans-fixed~\cite{gupta2021layout}}
        & \cg{\textbf{6.47}\std{6.8}} & \cg{0.133\std{33.3}} & \cg{17.1\std{4.3}} & \cg{\textbf{0.084}\std{6.5}} \\ %

        LayoutTrans~\cite{gupta2021layout} & 7.63\std{1.5} & \underline{0.068}\std{10.5} & \textbf{13.9}\std{1.1} & 0.127\std{6.9} \\

        MaskGIT$^\ast$~\cite{chang2022maskgit} & 52.1\std{6.2} & \textbf{0.015}\std{28.5} & 27.1\std{3.6} & \underline{0.101}\std{17.6} \\
        BLT~\cite{kong2022blt} & 88.2\std{22.6} & 1.030\std{43.9} & 116\std{31.4} & 0.153\std{26.3} \\

        BART~\cite{lewis2020bart} & 11.9\std{26.0} & 0.090\std{10.3} & 16.6\std{4.2} & 0.116\std{14.2} \\

        VQDiffusion$^\ast$~\cite{gu2022vector} & 7.46\std{6.5} & 0.178\std{46.4} & \underline{15.4}\std{2.3} & 0.193\std{13.8} \\

        LayoutDM & \underline{6.65}\std{6.2} & 0.162\std{11.6} & \textbf{13.9}\std{6.3} & 0.195\std{2.5} \\ \midrule
        Real data & 1.85 & 0.109 & 6.25 & 0.0214 \\
        \bottomrule
    \end{tabular}
\end{table}
\setlength\tabcolsep{6pt}
}

\subsection{Quantitative Evaluation} \label{sec:quantitative_evaluation}
\paragraph{C$\rightarrow$S+P, C+S$\rightarrow$P, Completion}
In these tasks, we inject conditions by masking.
We summarize comparisons in \cref{tab:comparison_inpainting}.
As task-specific models, we include LayoutVAE~\cite{jyothi2019layoutvae}, NDN-none~\cite{lee2020neural}, and LayoutGAN++~\cite{Kikuchi2021} for C$\rightarrow$S+P. We also adapt these models for C+S$\rightarrow$P.
LayoutDM outperforms other models except LayoutTrans~\cite{gupta2021layout} in completion.
The significant performance gap between LayoutDM and VQDiffusion* suggests the contribution of our proposals to go beyond the simple discrete diffusion models discussed in \cref{subsec:layout_diffusion_unconditional}.
Results in the completion suggest that a combination of padding and diffusion models is the primal key to the generation quality.
We find that FID and Maximum IoU are not highly correlated only in the completion task. We conjecture that Maximum IoU may become unstable when categories are also predicted, unlike the C$\rightarrow$S+P and C+S$\rightarrow$P tasks where categories are given.

\cref{fig:comparison_inpainting} shows the qualitative results of some models, including LayoutDM.
We can see that LayoutDM generates high-quality layouts with fewer layout aesthetics violations, such as misalignment and overlap, given diverse conditions.

\paragraph{Unconditional Generation}
\cref{tab:comparison_unconditional} summarizes the results of unconditional generation.
Unconditional layout generation methods often assume fixed order for element generation \eg top-to-bottom rather than random order for better generation quality by constraining the prediction.
For reference, we additionally report the results of LayoutTrans~\cite{gupta2021layout} trained on the fixed element order (LayoutTrans-fixed).
Although we design LayoutDM's primarily for conditional generation, LayoutDM achieves the best FID under random element order setting.
We conjecture that BLT's poor performance is due to train-test mask distribution inconsistency caused by their hierarchical masking strategy for training.
BLT masks a randomly sampled number of fields from a single semantic group \ie category, position, or size.
However, decoding starts with all masked tokens in inference.
The alignment metric of Real data stays at 0.109 in Rico.
Too small alignment values of LayoutTrans and MaskGIT can be a signal of producing trivial outputs in Rico.

{
\setlength{\tabcolsep}{1pt}
\begin{table}
\centering
\caption{
    Quantitative comparison in the refinement task.
    Top two results are highlighted in \textbf{bold} and \underline{underline}, respectively.
}
\label{tab:refinement}
\vspace*{-2mm}

    \begin{tabular}{@{}lcccccc@{}} \toprule
    \multicolumn{1}{r}{Dataset} & \multicolumn{3}{c}{Rico} & \multicolumn{3}{c}{PubLayNet} \\
    Model
    & FID $\downarrow$ & Max. $\uparrow$ & Sim $\uparrow$
    & FID $\downarrow$ & Max. $\uparrow$ & Sim $\uparrow$ \\
    \midrule

    \hspace{1mm}{\scriptsize Task-specific models} \\

    RUITE~\cite{rahman2021ruite}
    & \underline{3.23}\std{5.2} & \textbf{0.421}\std{0.8} & \textbf{0.221}\std{0.1} & \textbf{6.39}\std{1.1} & \textbf{0.415}\std{0.5} & \textbf{0.174}\std{0.3} \\

    \hspace{1mm}{\scriptsize Task-agnostic models} \\

    Noisy input
    & 134\std{0.2} & 0.213\std{0.2} & 0.177\std{0.1} & 130\std{0.2} & 0.242\std{0.1} & 0.147\std{0.1} \\

    LayoutDM
    & \textbf{2.77}\std{2.0} & \underline{0.370}\std{0.3} & \underline{0.205}\std{0.0} & \underline{6.75}\std{2.7} & \underline{0.352}\std{0.7} & \underline{0.149}\std{0.6} \\

    ~w/o logit adj.
    & 3.55\std{5.0} & 0.277\std{0.3} &0.168\std{0.2} & 7.95\std{1.7} & 0.310\std{0.3} & 0.127\std{0.2} \\

    \midrule

    Real data
    & 1.85 & 0.691 & 0.260 & 6.25 & 0.438 & 0.216 \\

    \bottomrule \end{tabular}
    \end{table}
    \setlength\tabcolsep{6pt}
}
{
    \setlength\tabcolsep{1pt}
    \vspace*{-2mm}
    \begin{figure}[t]
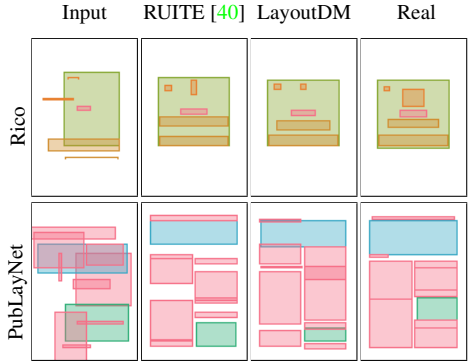

    \centering
    \newcommand{\length}{0.25}
    \newcommand{\labelmult}{2.5}
    \newcommand{\figcmd}[3]{
        \frame{\includegraphics[height=#3\hsize]{images/comparison/#1/refinement/#2_input.png}} &
        \frame{\includegraphics[height=#3\hsize]{images/comparison/#1/refinement/#2_ruite.png}} &
        \frame{\includegraphics[height=#3\hsize]{images/comparison/#1/refinement/#2_layoutdm.png}} &
        \frame{\includegraphics[height=#3\hsize]{images/comparison/#1/refinement/#2_gt.png}}
    }
    \footnotesize{
        \begin{tabular}{ccccc}
            & \multirow{2}{*}{Input} & \multirow{2}{*}{RUITE~\cite{rahman2021ruite}} & \multirow{2}{*}{\shortstack{LayoutDM}} & \multirow{2}{*}{Real} \\
            & & & & \\
            \raisebox{\labelmult\normalbaselineskip}[0pt][0pt]{\rotatebox[origin=c]{90}{Rico}} & \figcmd{rico25}{48504}{\length} \\
            \raisebox{\labelmult\normalbaselineskip}[0pt][0pt]{\rotatebox[origin=c]{90}{PubLayNet}} & \figcmd{publaynet}{PMC4963558_00004}{\length} \\
        \end{tabular}
    }
    \caption{Qualitative comparison in the refinement task.}
    \label{fig:comparison_refinement}
    \end{figure}
}

\paragraph{Refinement}
Our LayoutDM performs this task with a combination of the strong constraints of element categories, \ie, setting $\bm{z}^\mathrm{known} = \{(c_{1}, \texttt{[MASK]}, \ldots, \texttt{[MASK]}), \ldots\}$, and the weak constraints that geometric outputs appear near noisy inputs.
As an example of the weak constraint, we describe a constraint that imposes the x-coordinate estimate of $i$-th element close to the noisy continuous observation $\hat{x}_i$.
We denote a sliced vector of the prior term $\bm{\pi}$ in \cref{eq:prior_addition} that corresponds to the x-coordinate of $i$-th element as $\bm{\pi}_x^i \in \mathbb{R}^K$ and define by:
\begin{equation}
    \left[ \bm{\pi}_x^i \right]_j =
    \begin{cases}
        1 & \text{if}\ |\mathrm{loc}(j)-\hat{x}_{i}| < m~\text{and}~j \in X \\
        0 & \text{otherwise},
    \end{cases}
\end{equation}
where $m$ is a hyper-parameter indicating a margin,
$X$ is a set of indices denoting tokens for $x$ in the vocabularies,
and $\mathrm{loc}(j)$ is a function that returns the centroid value of $j$-th token in the vocabularies.
We define similar constraints for the other geometric variables and elements.

We summarize the performance in \cref{tab:refinement}.
We additionally report DocSim~\cite{patil2020read} (Sim) to measure the similarity of a predicted and its corresponding ground truth layout.
Imposing noisy geometric fields as a weak prior significantly improves the masking-only model and makes the performance much closer to RUITE~\cite{rahman2021ruite}, which is a denoising model not applicable to other layout tasks.
We compare some results in \cref{fig:comparison_refinement}.
Both LayoutDM and RUITE successfully recover complete layouts from non-trivially noisy layouts.

\paragraph{Relationship}
\begin{figure}[t]
    \centering
    \begin{minipage}{0.49\hsize}
        \adjustbox{trim={.025\width} 0 0 0,clip}{
            \includegraphics[width=1.05\hsize]{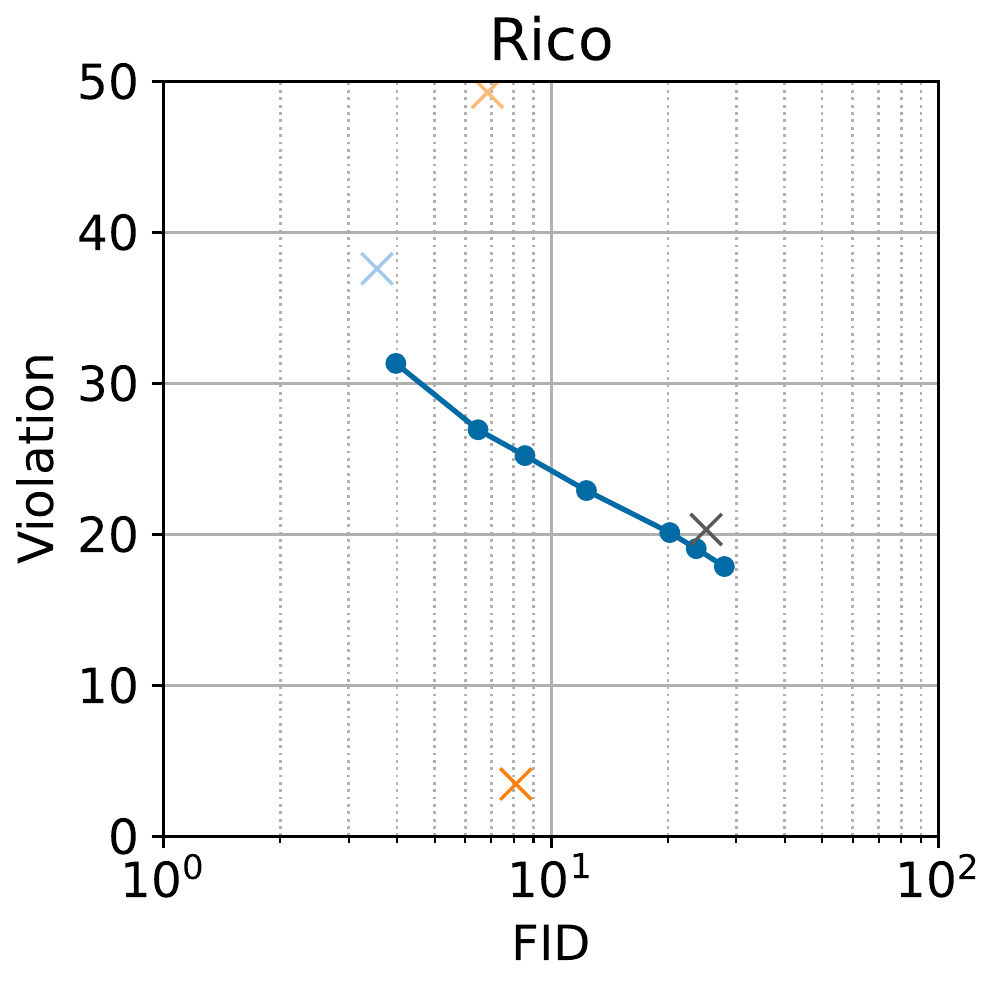}
        }
    \end{minipage}
    \begin{minipage}{0.49\hsize}
        \adjustbox{trim={.1\width} 0 0 0,clip}{
            \includegraphics[width=1.05\hsize]{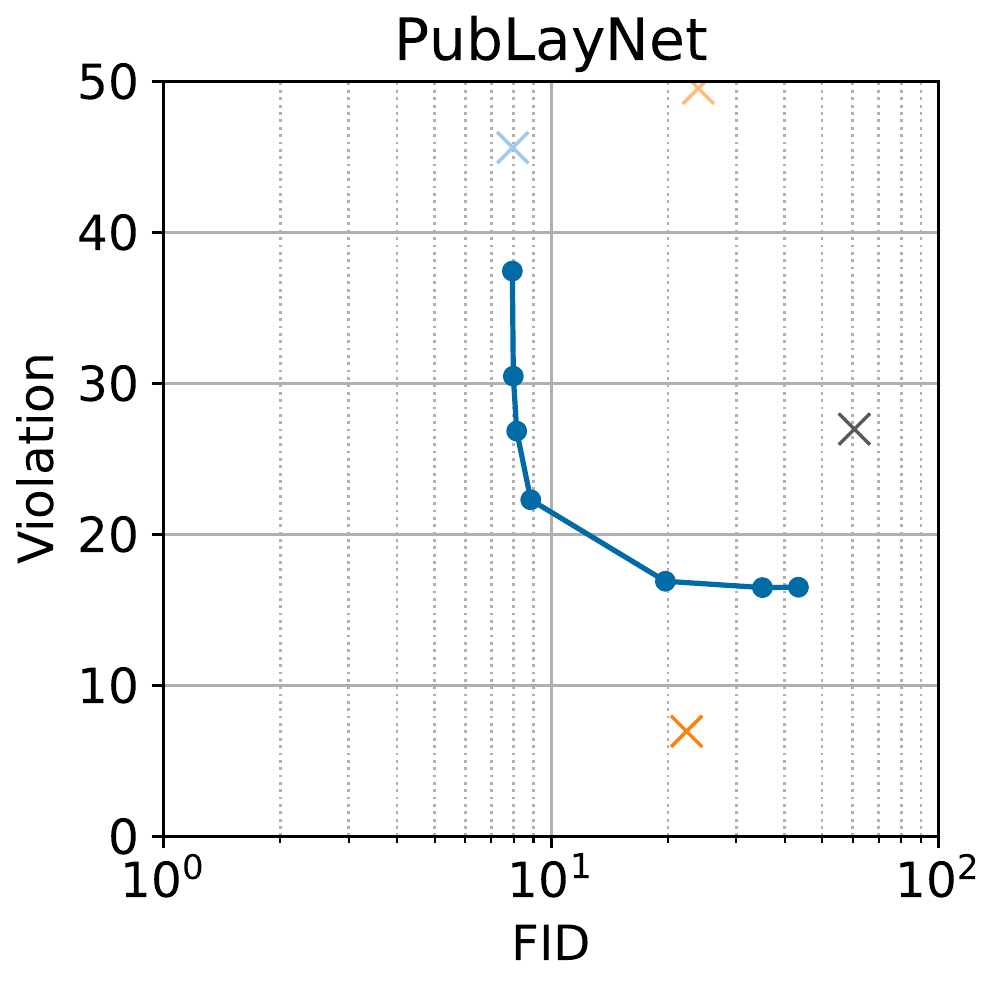}
        }
    \end{minipage}
    \adjustbox{trim={.05\width} {.175\height} {.025\width} {.175\height},clip}{
        \includegraphics[width=\linewidth]{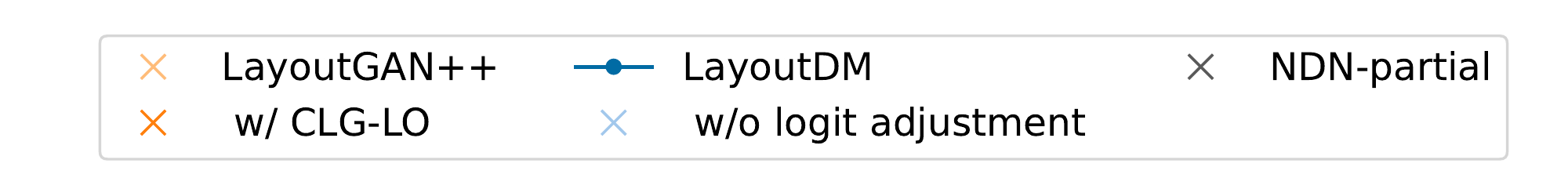}
    }
    \caption{
        Quality-violation trade-off in the relationship task.
        Lower scores indicate better performance for both metrics.
    }
    \label{fig:fid_violation_plots_relationship}
\end{figure}

{
    \setlength\tabcolsep{1pt}
    \begin{figure}[t]
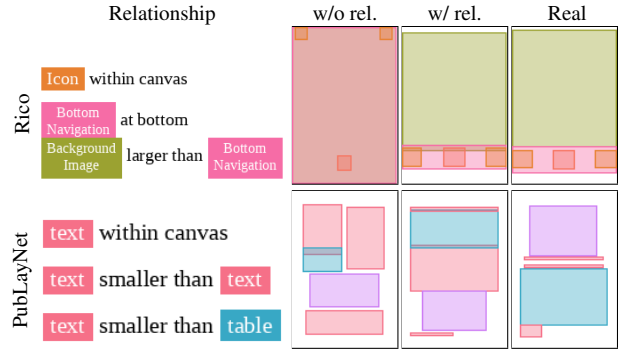

    \centering
    \newcommand{\length}{0.25}
    \newcommand{\labelmult}{2.5}
    \newcommand{\figcmd}[3]{
        \includegraphics[width=0.4\hsize]{images/comparison/#1/relationship/#2_relationship.png} &
        \frame{\includegraphics[height=#3\hsize]{images/comparison/#1/relationship/#2_unoptimized.png}} &
        \frame{\includegraphics[height=#3\hsize]{images/comparison/#1/relationship/#2_optimized.png}} &
        \frame{\includegraphics[height=#3\hsize]{images/comparison/#1/relationship/#2_gt.png}}
    }
    \footnotesize{
        \begin{tabular}{ccccc}
            & Relationship & w/o rel. & w/ rel. & Real \\
            \raisebox{\labelmult\normalbaselineskip}[0pt][0pt]{\rotatebox[origin=c]{90}{Rico}} & \figcmd{rico25}{36672}{\length} \\
            \raisebox{\labelmult\normalbaselineskip}[0pt][0pt]{\rotatebox[origin=c]{90}{PubLayNet}} & \figcmd{publaynet}{PMC3791796_00003}{\length} \\
        \end{tabular}
    }
    \caption{Qualitative comparison in the relationship task.}
    \label{fig:comparison_relationship}
    \end{figure}
}

\begin{figure}[t]
    \centering
    \begin{minipage}{0.49\linewidth}
        \adjustbox{trim={.05\width} {.025\height} 0 0,clip}{
            \includegraphics[width=1.05\linewidth]{images/time_fid_cwh_rico.pdf}
        }
    \end{minipage}
    \begin{minipage}{0.49\linewidth}
        \adjustbox{trim={.10\width} {.025\height} 0 0,clip}{
            \includegraphics[width=1.05\linewidth]{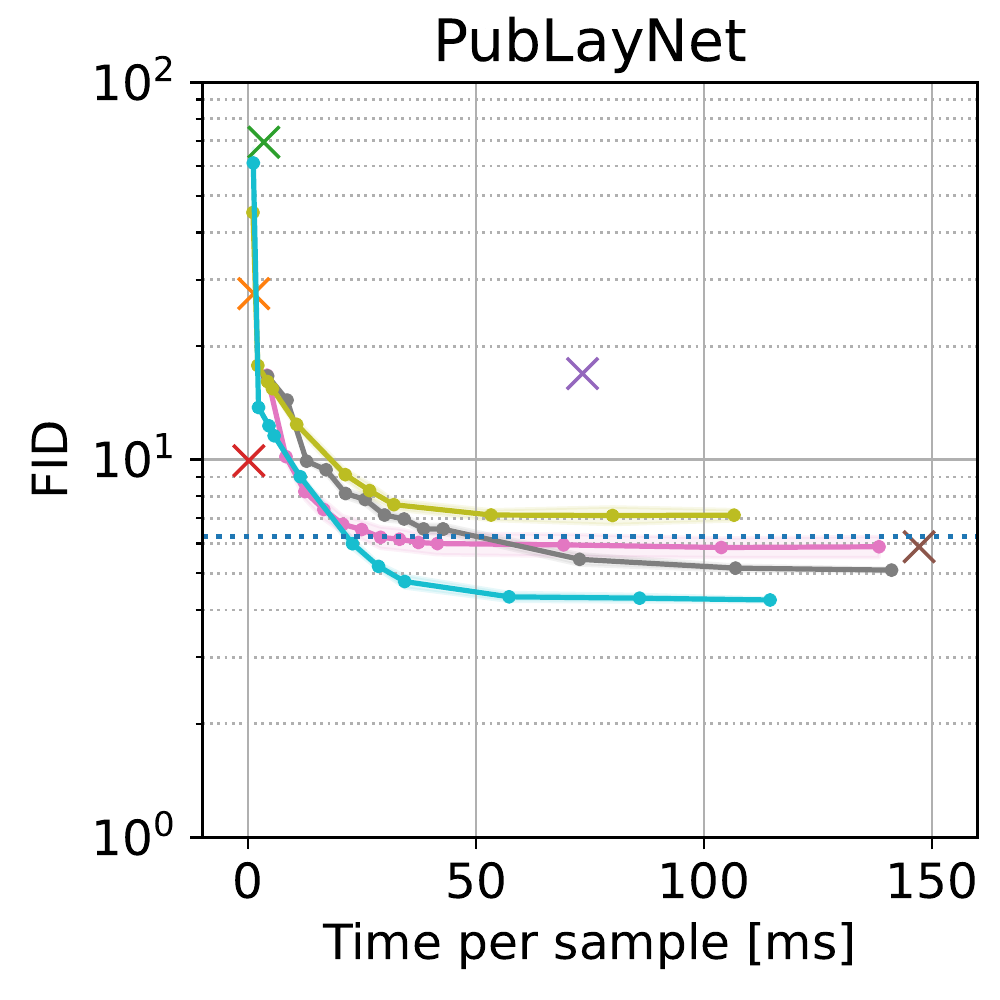}
        }
    \end{minipage}
    \adjustbox{trim={.10\width} 0 {.05\width} 0,clip}{
        \includegraphics[width=1.2\linewidth]{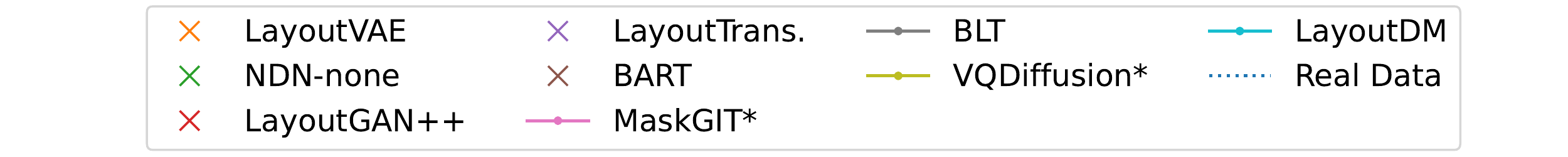}
    }
    \caption{Speed-quality trade-off of different models for C+S$\rightarrow$P.}
    \label{fig:tradeoff_time_fid}
\end{figure}
\begin{table}
    \centering
    {
        \setlength{\tabcolsep}{3pt}
        \caption{Ablation study results on layout-specific modification in unconditional generation of Rico~\cite{deka2017rico} dataset.}
        \label{tab:ablation_model_components}
        \begin{tabular}{lcc} \toprule
                                        & FID $\downarrow$          & Align. $\downarrow$         \\ \midrule
            LayoutDM                    & \textbf{6.65}\std{6.2}    & \underline{0.162}\std{11.6} \\
            ~w/o modality-wise diff.    & 7.32\std{7.0}             & \textbf{0.156}\std{10.1}    \\
            ~w/o decoupled pos. enc.    & \underline{6.78}\std{5.3} & 0.227\std{37.2}             \\
            ~w/ uniform-quantization    & 7.58\std{4.9}             & 0.256\std{32.2}             \\
            ~w/ percentile-quantization & 9.79\std{4.3}             & 0.232\std{20.0}             \\ \midrule
            Real Data                   & 1.85\std{0.0}             & 0.109\std{0.0}              \\
            \bottomrule
        \end{tabular}
        \setlength{\tabcolsep}{6pt}
    }
\end{table}

We use \cref{eq:prior_addition_by_gradient} to incorporate the relational constraints during the sampling step of LayoutDM.
We follow~\cite{Kikuchi2021} to employ the loss functions penalizing size and location relationships between elements that do not match user specifications.
We define the loss functions for continuous bounding boxes, and we have to convert the predicted discrete bounding boxes to continuous ones in a differentiable manner.
Given estimated probabilities of discrete x-coordinates $p(x)$, for example, we compute the continuous x-coordinate $\bar{x}$ by %
$\bar{x} = \sum_{n \in X} p(x\!=\!n) \, \mathrm{loc}(n)$.
Similar conversion applies to the other attributes.
Empirically, we find that
applying the logit adjustment multiple times (three times in our experiments)
to each diffusion step moderately improves performance.

We compare LayoutDM with two task-specific approaches: NDN-partial~\cite{lee2020neural} and CLG-LO based on LayoutGAN++~\cite{Kikuchi2021}.
We show the results in \cref{fig:fid_violation_plots_relationship}.
We additionally report constraint violation error rates~\cite{Kikuchi2021}.
LayoutDM can control the strength of the logit adjustment as in \cref{eq:prior_addition} and produces an FID-violation trade-off curve.
LayoutDM is comparable to NDN-partial in Rico and outperforms NDN-partial by a large margin in PubLayNet.
Although LayoutDM is inferior to CLG-LO in both datasets, note that the average runtime of CLG-LO is 4.0s, which is much slower than 0.5s in LayoutDM.
We show some results of LayoutDM in \cref{fig:comparison_relationship}.

\subsection{Speed-Quality Trade-off}
Runtime is also essential for a controllable generation.
We show a speed-quality trade-off curve for C+S$\rightarrow$P as shown in \cref{fig:tradeoff_time_fid}.
The Transformer encoder-only models, such as LayoutDM and BLT, can achieve fast generation at the sacrifice of quality.
We employ fast-sampling strategy employed in discrete diffusion models~\cite{austin2021structured} for LayoutDM by $p_{\theta}(\bm{z}_{t-\Delta}|\bm{z}_{t}) \propto \sum_{\tilde{\bm{z}}_{0}} q(\bm{z}_{t-\Delta},\bm{z}_{t}|\tilde{\bm{z}}_{0}) \tilde{p}_{\theta}(\tilde{\bm{z}}_{0}|\bm{z}_{t})$, where $\Delta \in \mathbb{N}$ indicates a step size for generation in $\frac{T}{\Delta}$ steps.
Despite being a task-agnostic model, LayoutDM achieves the best quality-speed trade-off except for task-specific LayoutGAN++~\cite{Kikuchi2021} that runs under 10ms.

\subsection{Ablation Study}  \label{sec:ablation_study}
We investigate whether techniques in \cref{subsec:layout_diffusion_unconditional} improve the performance.
First, we evaluate a choice of quantization methods for the geometric fields of elements.
Instead of KMeans, we compute centroids for the quantization by:
\begin{itemize}[noitemsep,nolistsep,leftmargin=*]
    \item Uniform: This is dataset-agnostic quantization, which is popular in previous works~\cite{arroyo2021variational,gupta2021layout,kong2022blt}. Following~\cite{gupta2021layout}, we choose $\{0.0, \frac{1}{B}, \ldots, \frac{B - 1}{B}\}$ and $\{\frac{1}{B}, \ldots, \frac{B - 1}{B}, 1.0\}$ for the position and size, respectively.
    \item Percentile: we sort the data into equally sized groups and obtain average values for each group as the centroids. This is dataset-specific quantization similar to KMeans.
\end{itemize}
We show the result at the bottom of \cref{tab:ablation_model_components}.
We additionally report the Alignment metric (Align.) used in~\cite{Kikuchi2021} since the choice of the quantization affects the alignment between elements.
Compared to Linear and Percentile, KMeans quantization significantly improves both FID and Alignment.
We confirm that our modality-wise diffusion and decoupled positional encoding both moderately improve the performance, as we show at the top half of \cref{tab:ablation_model_components}.
\section{Discussion}
LayoutDM is based on diffusion models for discrete state-space.
Using continuous state-space as in latent diffusion models~\cite{rombach2022high} would be interesting. Extension of LayoutDM to handle various layout properties such as color~\cite{kikuchi2023generative} and image/text~\cite{yamaguchi2021canvasvae} is also appealing.

We believe our proposed logit adjustment can incorporate more attributes.
Attribute-conditional LayoutGAN~\cite{li2020attribute} considers area, aspect ratio, and reading order of elements for fine-grained control. Since these attributes can be easily converted to the size and location relationship constraints, incorporating them with our LayoutDM is not very difficult.

\paragraph{Potential negative impact}
Our model might be used to automatically generate the basic structure of fake websites or mobile applications, which could lead to scams or the spreading of misinformation.

{\small
\bibliographystyle{ieee_fullname}
\bibliography{egbib}
}

\clearpage
\appendix

\section{Implementation Details}

\subsection{Baseline}
We explain more details on task-agnostic layout generation baselines using masking, especially when the original model is not designed for layout generation.
We mostly describe unconditional generation cases, but partial layout fields can be easily injected by hard masking.

\noindent \textbf{BART}: BART is a denoising autoencoder and was originally designed for learning a sequence-to-sequence model for text generation.
Text is usually represented as a 1D sequence of discrete tokens.
Since we also handle the shuffled layout as a 1D sequence of discrete tokens during training, a BART-like model may be another solid baseline.
To build a task-agnostic layout generation model, we apply random masking similar to the noise pattern of MaskGIT~\cite{chang2022maskgit}, instead of text-specific noises, such as span-level masking.

\noindent \textbf{MaskGIT$^\ast$}:
MaskGIT~\cite{chang2022maskgit} is originally built for unconditional image generation.
Following recent two-stage approaches for efficient image modeling, such as VQ-GAN~\cite{esser2021taming}, MaskGIT first generates a small number of discrete tokens and subsequently decodes those tokens into a continuous high-dimensional image by a pre-trained neural decoder.
We consider the first generation part of MaskGIT to be another baseline.
We use \texttt{[PAD]} to enable variable-length generation.
For a masking schedule during decoding, \ie fraction of the tokens masked in each iteration, we employ a cosine schedule as in MaskGIT.

\noindent \textbf{VQDiffusion$^\ast$}: VQDiffusion~\cite{gu2022vector} is a discrete diffusion-based model designed for text-to-image generation.
To adapt VQDiffusion for conditional layout generation with minimal modification, we (i) remove the text conditioning branch in the reverse process, (ii) replace the image tokens with layout tokens, and (iii) add \texttt{[PAD]} token to enable variable-length generation. As described in the main manuscript, there are three major differences between VQDiffusion$^\ast$ and our proposed LayoutDM: modality-wise diffusion, decoupled positional encoding, and adaptive quantization.

We adjust the number of parameters for each model to have about 12M parameters for a fair comparison.
We show the exact numbers in \cref{tab:number_of_parameters}.

\subsection{Relationship Guidance}
Similarly to the main manuscript, let us denote the predicted coordinates of an $i$-th element $(\hat{x}_{i},\hat{y}_{i},\hat{w}_{i},\hat{h}_{i}) \in [0, 1]^{4}$. We follow ~\cite{Kikuchi2021} to define the loss for penalizing size and location relationships between elements that do not match user specifications. For example, if we want to make the $j$-th element larger than the $i$-th element, the loss is defined by:
\begin{equation}
    g_{lg}(i, j) = \max\left(\left(1+\gamma\right)\hat{w}_{i}\hat{h}_{i}-\hat{w}_{j}\hat{h}_{j}, 0\right),
\end{equation}
where $\gamma$ is a tolerance parameter, which is empirically set to 0.1.
If we want to make the $j$-th element above the $i$-th element, the loss is defined by:
\begin{equation}
    g_{ab}(i, j) = \max\left(\left(\hat{y}_{j} + \frac{\hat{h}_{j}}{2}\right) - \left(\hat{y}_{i} - \frac{\hat{h}_{i}}{2}\right), 0\right),
\end{equation}
which compares the bottom of the $j$-th element and the top of the $i$-th element.
Please refer to the code for losses for the rest of the relationships.

Although it is not experimentally demonstrated, we believe that it is also possible to incorporate area, aspect ratio, and reading order constraints used in Attribute-conditioned GAN~\cite{li2020attribute}.
\begin{itemize}[noitemsep,nolistsep,leftmargin=*]
\item Area: given a target area of the element $a_{i} \in \mathbb{R}$, we use $|a_{i} - \hat{h}_{i}\hat{w}_{i}|$ as a loss.
\item Aspect ratio: Given a target aspect ratio $r_{i} \in \mathbb{R}$, we use $|r_{i} - \frac{\hat{h}_{i}}{\hat{w}_{i}}|$ as a loss.
\item Reading order: we follow [22] and define that the reading order solely depends on the distance between the left-top of the canvas and each element.
We first compute the distance by $\hat{d}_{i}=\sqrt{(\hat{x}_{i} - \frac{\hat{w}_{i}}{2})^{2} + (\hat{y}_{i} - \frac{\hat{h}_{i}}{2})^{2}}$.
We can use $\max(\hat{d}_{i} - \hat{d}_{j}, 0)$ as a loss to make the $i$-th element come before the $j$-th element in the reading order.
\end{itemize}

\subsection{Hyper-parameters}
We search for the best hyper-parameters using a validation set.
During sampling from $p_{\theta}\!\left(\bm{z}_{t-1}|\bm{z}_{t}\right)$ for all the tasks, we search for $p$ used in nucleus (or top-$p$) sampling~\cite{holtzman2019curious} out of $\{0.90, 0.95, 0.99, 1.0\}$.
We train the models for 50 and 20 epochs in Rico and PubLayNet, respectively.

We attempt a grid search for additional hyper-parameters in the refinement task.
The ranges of possible values are the following: the distance margin $m$ in $\{0.1, 0.2\}$ and the weighting term $\lambda_{\pi}$ in $\{1.0, 2.0, 3.0, 4.0, 5.0\}$.

\begin{table}[t]
    \centering
    \begin{tabular}{ccc} \toprule
        & Rico & PubLayNet \\ \midrule
        LayoutVAE~\cite{jyothi2019layoutvae}~(C$\rightarrow$S+P) & 13.2 & 13.0 \\
        NDN-none~\cite{lee2020neural}~(C$\rightarrow$S+P) & 21.8 & 21.8 \\
        LayoutGAN++~\cite{Kikuchi2021}~(C$\rightarrow$S+P) & 12.9 &  12.9 \\
        LayoutVAE~\cite{jyothi2019layoutvae}~(C+S$\rightarrow$P) & 14.7 & 14.5 \\
        NDN-none~\cite{lee2020neural}~(C+S$\rightarrow$P) & 14.8 & 14.8  \\
        LayoutGAN++~\cite{Kikuchi2021}~(C+S$\rightarrow$P) & 13.0 & 13.0 \\
        LayoutTrans~\cite{gupta2021layout} & 12.7 & 12.7 \\
        LayoutTrans-fixed~\cite{gupta2021layout} & 12.7 & 12.7 \\
        MaskGIT*~\cite{chang2022maskgit} & 12.7 & 12.7 \\
        BLT~\cite{kong2022blt} & 12.7 & 12.7 \\
        RUITE~\cite{rahman2021ruite} & 12.7 & 12.7 \\
        BART~\cite{lewis2020bart} & 12.8 & 12.8 \\
        VQDiffusion*~\cite{gu2022vector} & 12.4 & 12.4 \\
        LayoutDM & 12.4 & 12.4 \\ \bottomrule
    \end{tabular}
    \caption{
        The number of parameters [M] used for each model.
    }
    \label{tab:number_of_parameters}
\end{table}

\subsection{Evaluation}
In unconditional generation, the model generates 1,000 samples from the random seed.
In conditional generation, the test set of each dataset is used to make a partial input for conditional generation and the model generates one sample per each data in the test set.

\section{Additional Results}

\subsection{Ablation Study}
\paragraph{State space}
Continuous state space diffusion models have gained much attention compared to discrete state space models.
Recently, Li~\etal~\cite{li2022diffusion} propose DiffusionLM that adapts the continuous models to handle discrete text generation.
DiffusionLM introduces an embedding and rounding step to bridge the continuous and discrete state spaces.
We train DiffusionLM (with 12.6M parameters) and show the results in \cref{tab:state-space}.
We show the results of DiffusionLM with embedding dimensions $d=16$ because it works best out of $\{16, 64, 128\}$ in Rico~\cite{deka2017rico} dataset.
Although We tried different samplers (DDPM~\cite{ho2020denoising} and DDIM~\cite{song2020denoising}) and training timesteps, DiffusionLM is still far behind the discrete state space models in layout generation as shown in \cref{tab:state-space}.

\begin{table}[t]
    \centering
    \begin{tabular}{ccccc} \toprule
        & State & \#steps & Sampler & FID $\downarrow$ \\ \midrule
        LayoutDM & dis. & 100 & - & \textbf{6.65} \\
        VQDiffusion*~\cite{gu2022vector} & dis. & 100 & - & \underline{7.46} \\
        \multirow{4}{*}{DiffusionLM~\cite{li2022diffusion}} & con. & 100 & DDIM & 34.5 \\
         & con. & 100 & DDPM & 24.8 \\
         & con. & 1000 & DDIM & 33.8 \\
         & con. & 1000 & DDPM & 22.8 \\ \bottomrule
    \end{tabular}
    \caption{
        Ablation study results on the choice of state spaces: discrete (dis.) and continuous (con.), in the unconditional generation task of Rico~\cite{deka2017rico} dataset. Top two results are highlighted in \textbf{bold} and \underline{underline}, respectively.
    }
    \label{tab:state-space}
\end{table}

\paragraph{Refinement}
The logit adjustment proposed in the main manuscript has some choices for injecting positional prior.
Without loss of generality, we describe a constraint that imposes the x-coordinate estimate of $i$-th element close to the noisy continuous observation $\hat{x}_i$.
We denote a sliced vector of the prior term $\pi\!\left(\bm{z}_{t-1}\right)$ that corresponds to the x-coordinate of $i$-th element as $\bm{\pi}_x^i \in \mathbb{R}^K$.

\begin{itemize}[noitemsep,nolistsep,leftmargin=*]
    \item Gaussian: $j$-th token is more likely to be sampled when $\mathrm{loc}(j)$ is closer to $\hat{x}_{i}$. The prior is defined by:
    \begin{equation}
        \begin{aligned}
        & \left[ \bm{\pi}_x^i \right]_j = \\
        & \begin{cases}
            \left(\mathrm{loc}\left(j\right)-\hat{x}_{i}\right)^{2} & \text{if}\ \left|\mathrm{loc}(j)-\hat{x}_{i}\right| < m~\text{and}~j \in X \\
            0 & \text{otherwise},
        \end{cases}
        \end{aligned}
    \end{equation}
    The ranges of possible values are similar to our method used in the main manuscript (Default).

    \item Negation: $j$-th token is never sampled when $\mathrm{loc}(j)$ is far away from $\hat{x}_{i}$. The prior is defined by:
    \begin{equation}
        \begin{aligned}
        \left[ \bm{\pi}_x^i \right]_j =
        \begin{cases}
            0 & \text{if}\ \left|\mathrm{loc}(j)-\hat{x}_{i}\right| < m~\text{and}~j \in X \\
            -\infty & \text{otherwise}.
        \end{cases}
        \end{aligned}
    \end{equation}
    The ranges of possible values are the following: the distance margin $m$ in $\{0.2, 0.4, 0.6, 0.7, 0.8, 0.9\}$.
\end{itemize}

We show the quantitative evaluation results in \cref{tab:ablation_refinement}.
We can see that Default outperforms other possible choices by a large margin.

\begin{table}[t]
    \centering
    \begin{tabular}{cccc} \toprule
        & FID $\downarrow$ & Max. $\uparrow$ & Sim $\uparrow$ \\ \midrule
        Default & \textbf{2.77}\std{2.0} & \textbf{0.370}\std{0.3} & \textbf{0.205}\std{0.0} \\
        Gaussian & 5.82\std{1.7} & \underline{0.330}\std{0.4} & \underline{0.188}\std{0.3} \\ %
        Negation & \underline{3.78}\std{3.1} & 0.276\std{0.4} & 0.169\std{0.3} \\ \bottomrule %
    \end{tabular}
    \caption{
    Ablation study results on the choice of logit adjustment methods in the refinement task. Top two
results are highlighted in \textbf{bold} and \underline{underline}, respectively.
    }
    \label{tab:ablation_refinement}
\end{table}

\subsection{Speed-Quality Trade-off}
\label{subsec:speed_quality_tradeoff}
We show more speed-quality trade-off curves in \cref{fig:sup_tradeoff_time_fid}.
We perform generation with a batch size of 64 and report the average runtime to generate a single layout for all the models.
Lightly colored regions around the line plots, such as the one in BLT for C$\rightarrow$S+P in Rico represent the standard deviation of three trials for each model, though the deviations are too small to see in most cases.

\subsection{More Results}
We show more results compared with task-specific baselines in C$\rightarrow$S+P (\cref{fig:sup_comparison_c_publaynet}), C+S$\rightarrow$P (\cref{fig:sup_comparison_cwh_publaynet}), unconditional generation (\cref{fig:sup_comparison_unconditional_publaynet}), the refinement task (\cref{fig:sup_comparison_refinement_publaynet}) for PubLayNet.
Typical failure cases are frequent overlap between elements (often in BLT), unnecessarily broad blank space (often in LayoutTrans.), and lack of diversity.
We show more results in C$\rightarrow$S+P (\cref{fig:sup_comparison_c_rico}), C+S$\rightarrow$P (\cref{fig:sup_comparison_cwh_rico}), unconditional generation (\cref{fig:sup_comparison_unconditional_rico}), the refinement task (\cref{fig:sup_comparison_refinement_rico}) for Rico.
Rico is more difficult to generate since the number of categories is large and elements are less aligned compared to PubLayNet.

\subsection{Diversity-Fidelity Trade-off}
We introduce density and coverage metrics by~\cite{naeem2020reliable} to analyze the results from a different viewpoint.
Density measures fidelity; \ie, how closely generated samples resemble real ones.
Coverage measures diversity; \ie, whether generated samples cover the full variability of the real samples.
We plot the diversity and fidelity of iterative refinement-based models in \cref{fig:sup_tradeoff_density_coverage} as we increase the number of timesteps for the iterative prediction.
Discrete diffusion-based models usually have higher coverage scores and lower density scores.
We conjecture that the coverage difference comes from the inference decoding strategy.
BLT~\cite{kong2022blt} and MaskGIT$^\ast$~\cite{chang2022maskgit} fix high-confident predictions and re-initialize lower-confident fields by \texttt{[MASK]} for the next step that leads to higher fidelity.
In contrast, discrete diffusion-based models \textit{randomly} corrupt the predictions and result in higher diversity.

\subsection{Alignment and Overlap}
We additionally show the metrics reported in many previous works: Alignment and Overlap.
Note that these metrics only capture the fidelity of generated layouts.
There are a few variants for both Alignment~\cite{li2019layoutgan,lee2020neural,li2020attribute,Kikuchi2021} and Overlap~\cite{li2019layoutgan,li2020attribute,Kikuchi2021}.
We employ the definition in \cite{Kikuchi2021}.
We scale the values of Alignment by $100\times$ for visibility.
For reference, we show Alignment and Overlap computed in a validation set as~\emph{Real data}.
The lowest score in Alignment or Overlap does not always mean the best performance for a model, but a model closest to~\emph{Real data} is the best model.
We show the result in \cref{fig:sup_overlap_alignment}.
In the fixed-length generation \ie C$\rightarrow$S+P and C+S$\rightarrow$P, LayoutDM performs almost comparably to VQDiffusion*~\cite{gu2022vector} and BART~\cite{lewis2020bart}, and better than the other models.
In the variable-length generation \ie the completion task and unconditional generation, autoregressive models, such as BART~\cite{lewis2020bart} and LayoutTrans.~\cite{gupta2021layout}, are moderately better than LayoutDM.
This result is reasonable since these models predict the fields one by one.
Diffusion-based models, such as LayoutDM and VQDiffusion$^\ast$, are better than BLT and MaskGIT$^\ast$.
We believe this is because diffusion models avoid the error accumulation in iterative prediction according to~\cite{gu2022vector}.

\newcommand{\tflength}{0.475}
\newcommand{\tffig}[1]{
    \adjustbox{trim={.025\width} {.025\height} {.025\width} 0,clip}{
        \includegraphics[width=\tflength\linewidth]{images/supplementary/time_fid/#1.pdf}
    }
}
\begin{figure*}[t]
    \centering
    \begin{tabular}{cc}
        \multicolumn{2}{c}{C$\rightarrow$S+P} \\
        \tffig{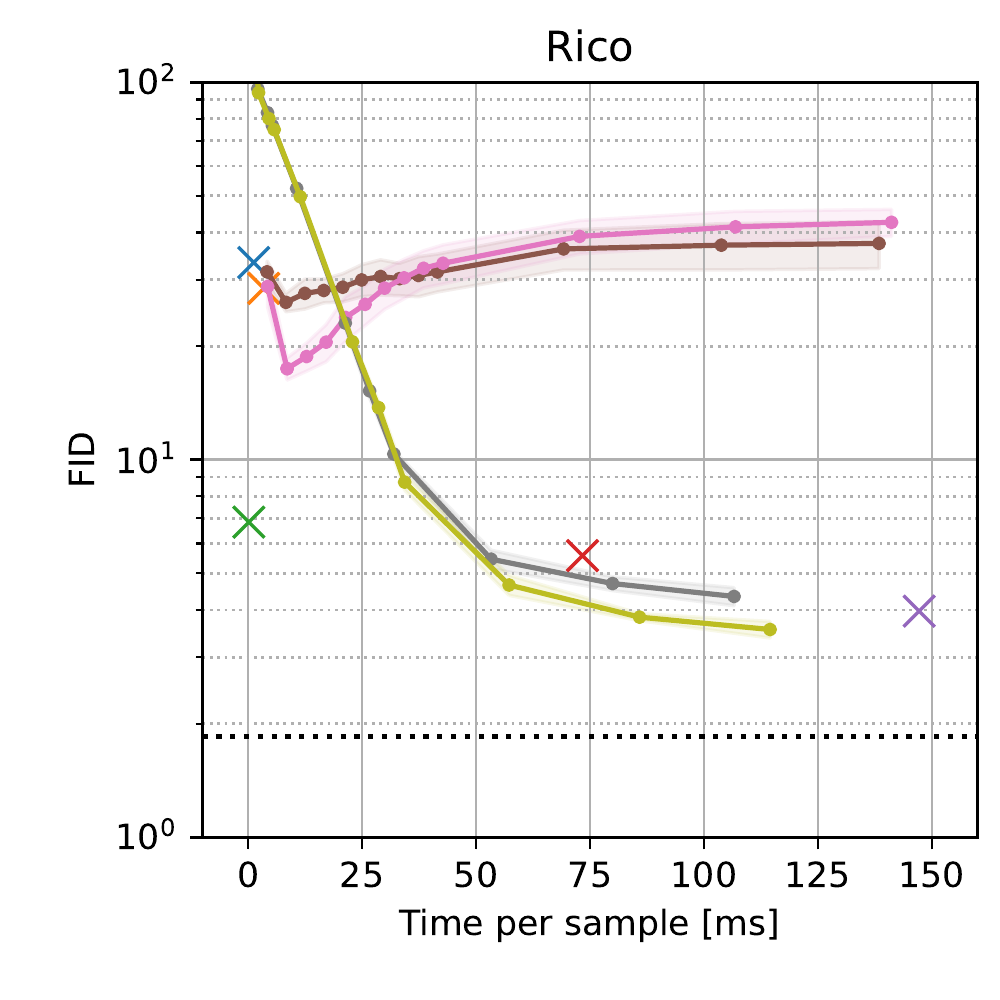} & \tffig{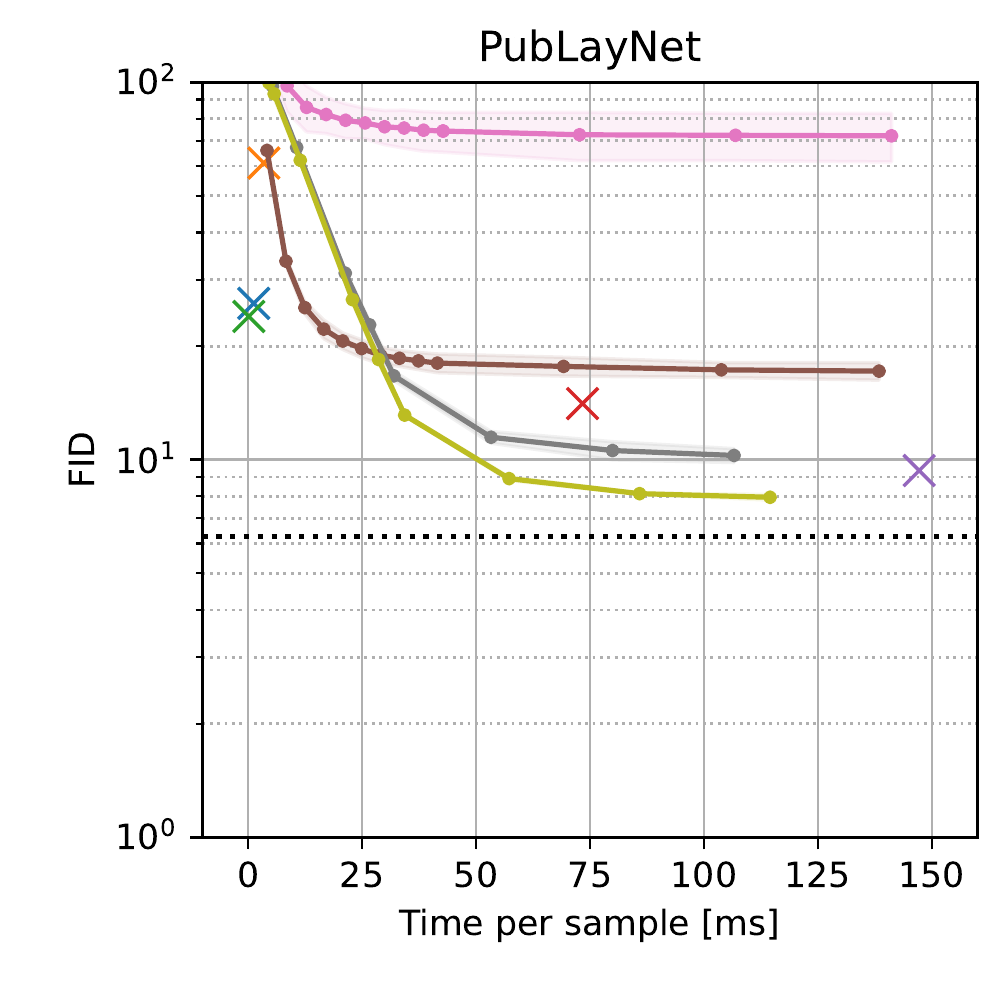} \\
        \multicolumn{2}{c}{
            \adjustbox{trim=0 0 0 0,clip}{
                \includegraphics[width=\linewidth]{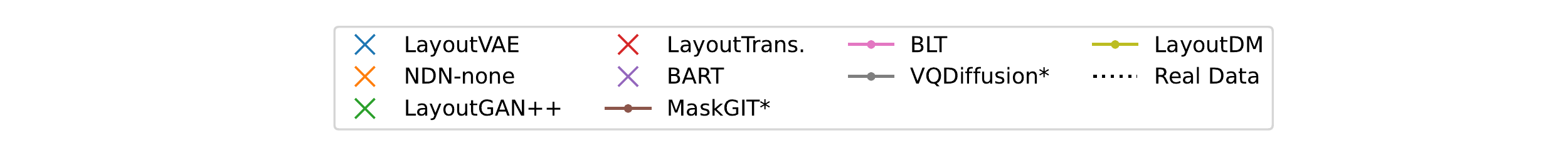}
            }
        } \\
        \multicolumn{2}{c}{C+S$\rightarrow$P} \\
        \tffig{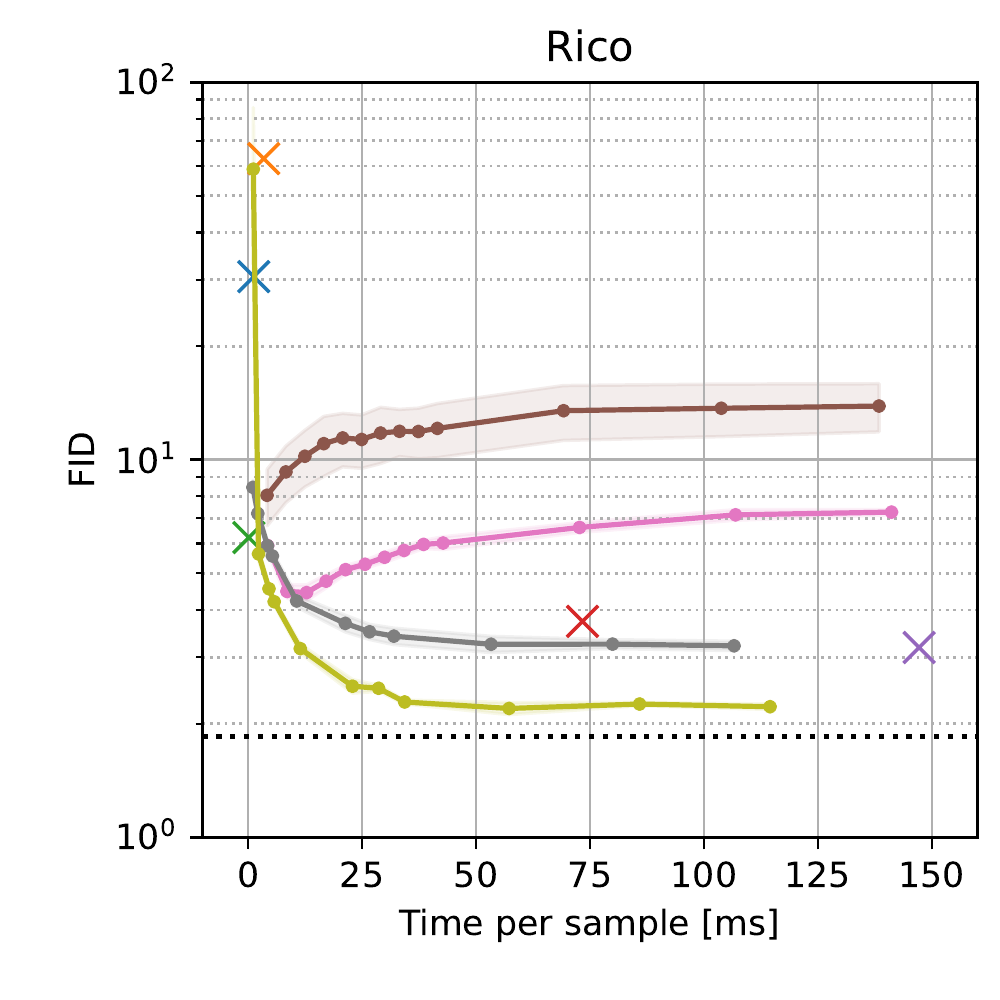} & \tffig{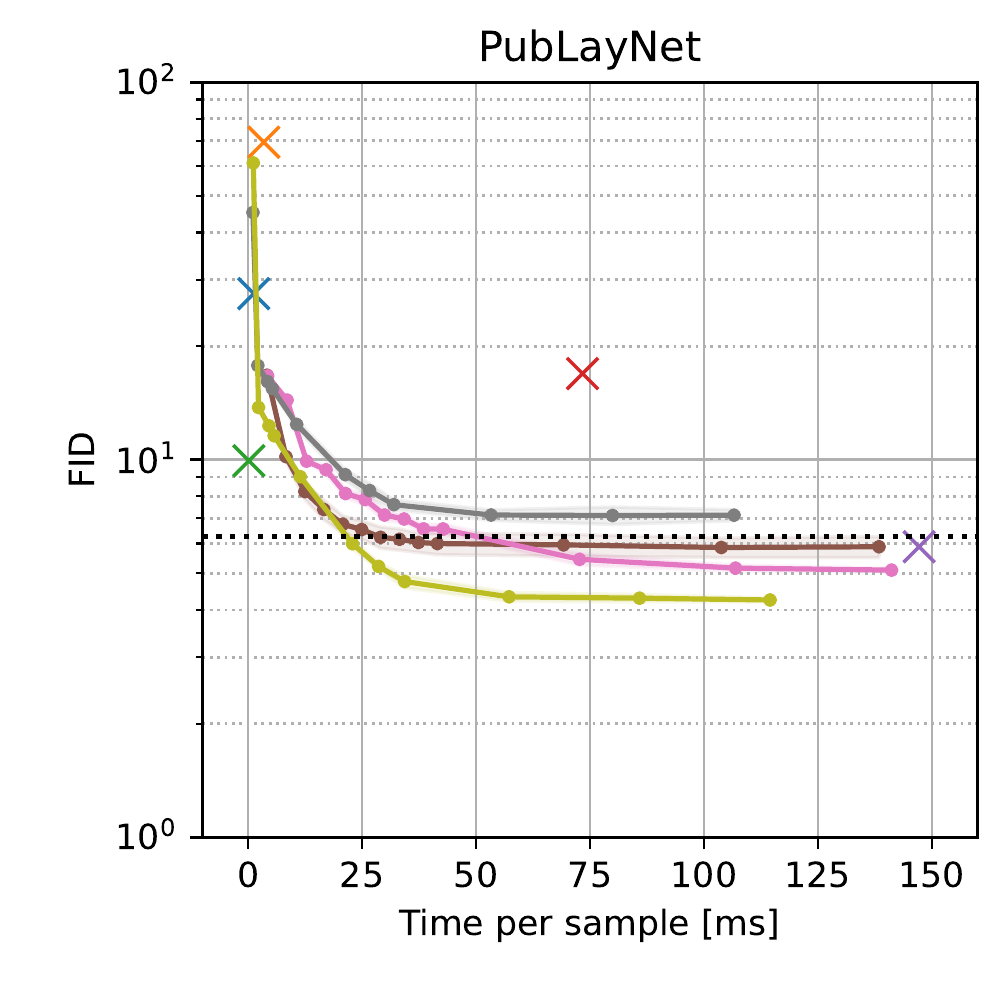} \\
        \multicolumn{2}{c}{
            \adjustbox{trim=0 0 0 0,clip}{
                \includegraphics[width=\linewidth]{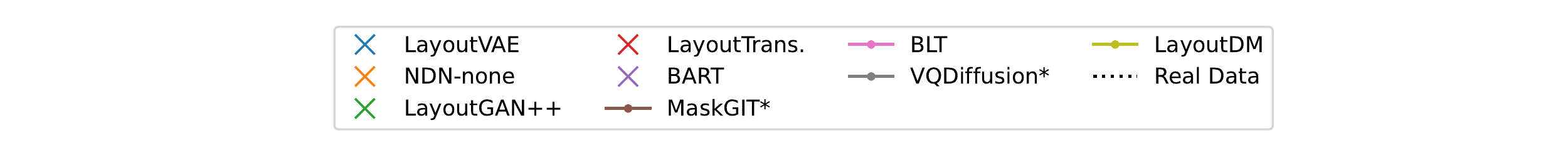}
            }
        }
    \end{tabular}
    \caption{Speed-quality trade-off of different models.}
    \label{fig:sup_tradeoff_time_fid}
\end{figure*}
\begin{figure*}[t]
    \addtocounter{figure}{-1}
    \centering
    \begin{tabular}{cc}
        \multicolumn{2}{c}{Partial} \\
        \tffig{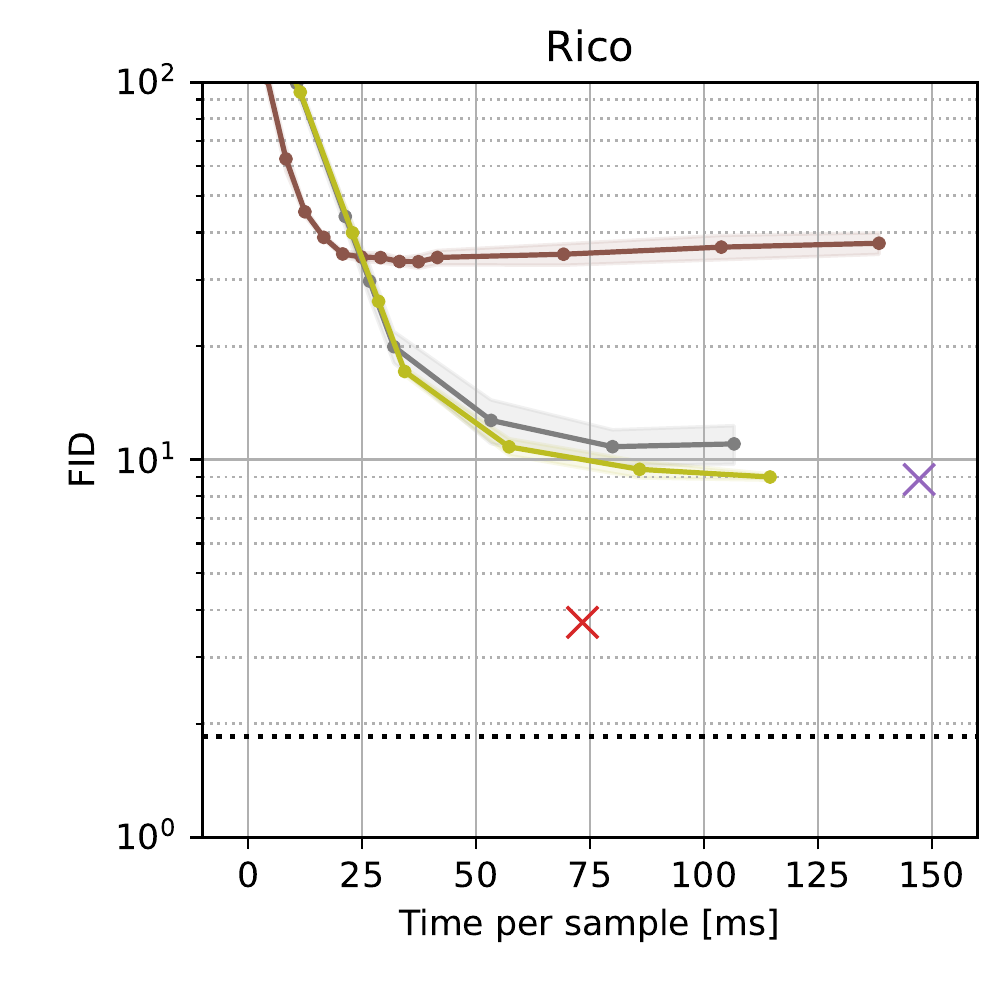} & \tffig{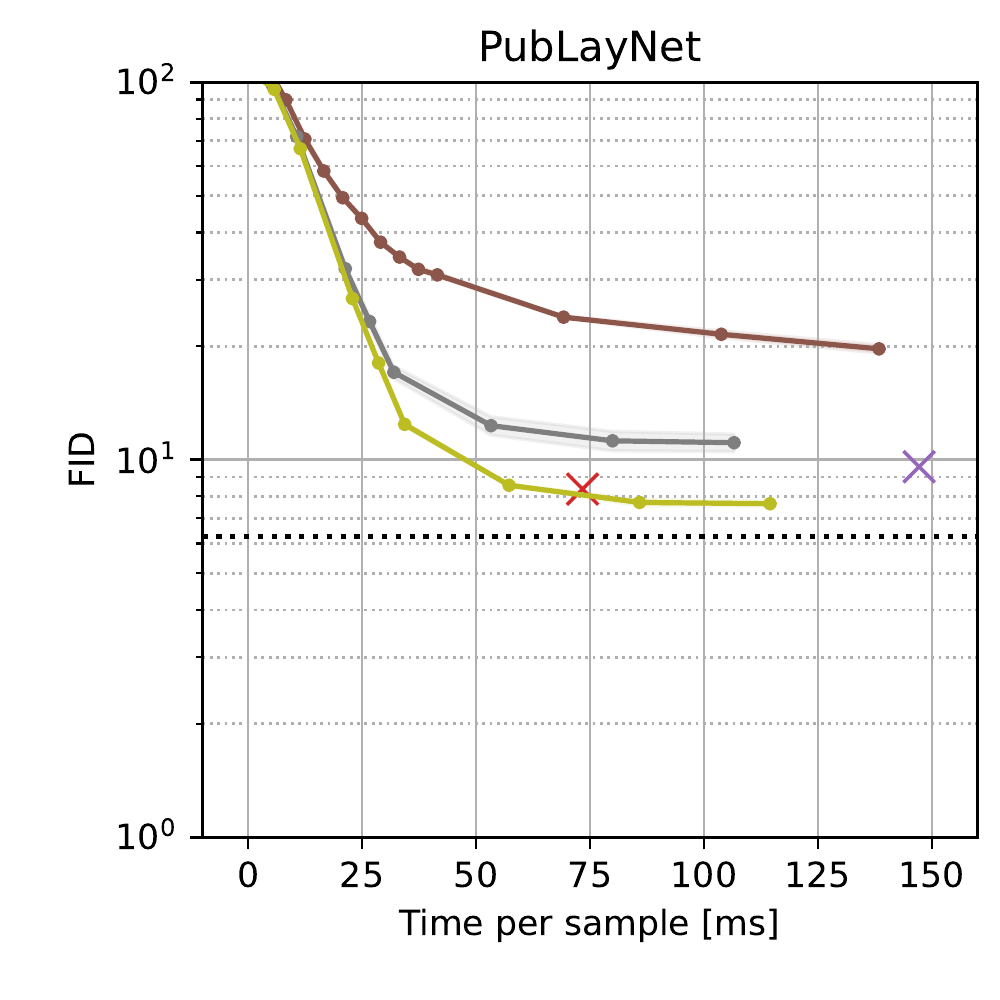} \\
        \multicolumn{2}{c}{
            \adjustbox{trim=0 0 0 0,clip}{
                \includegraphics[width=\linewidth]{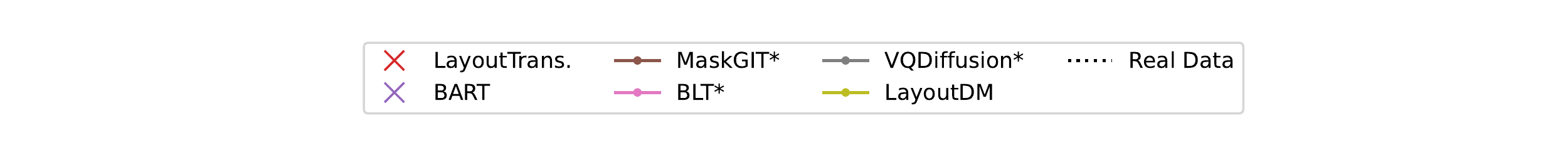}
            }
        } \\
        \multicolumn{2}{c}{Unconditional} \\
        \tffig{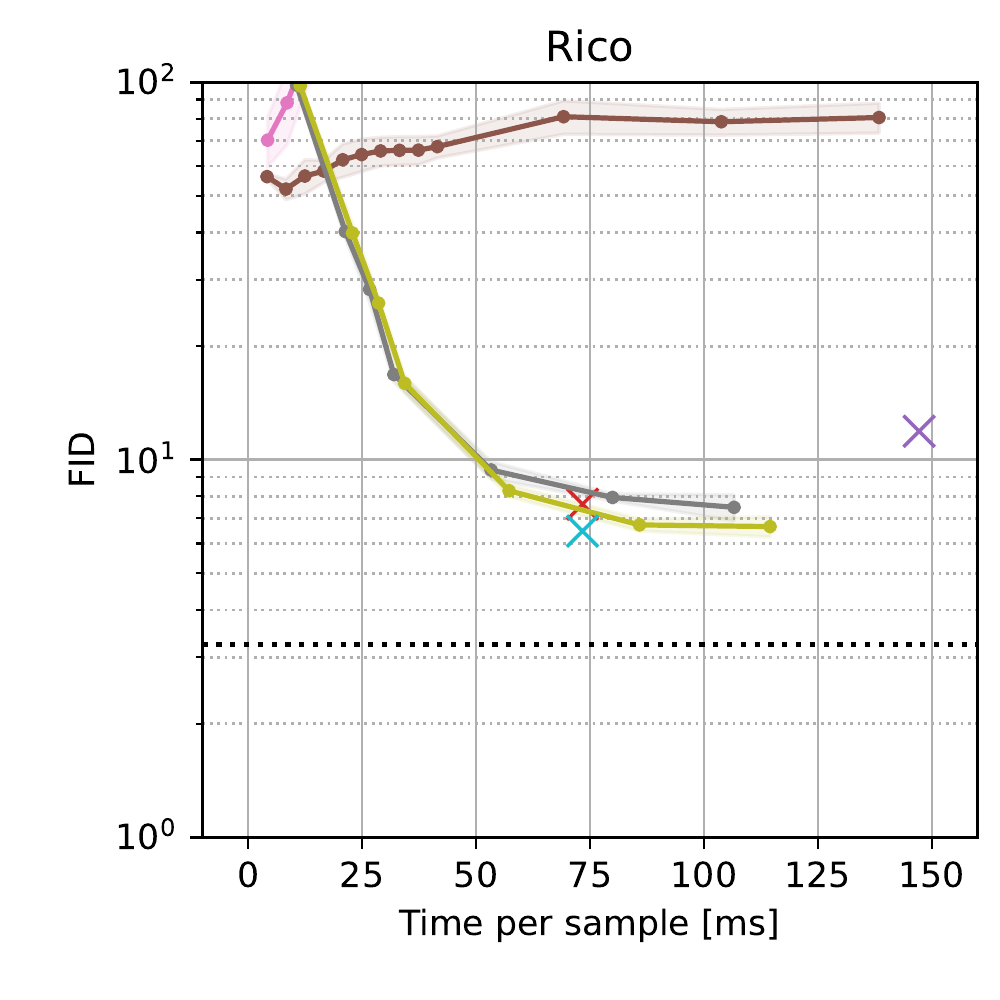} & \tffig{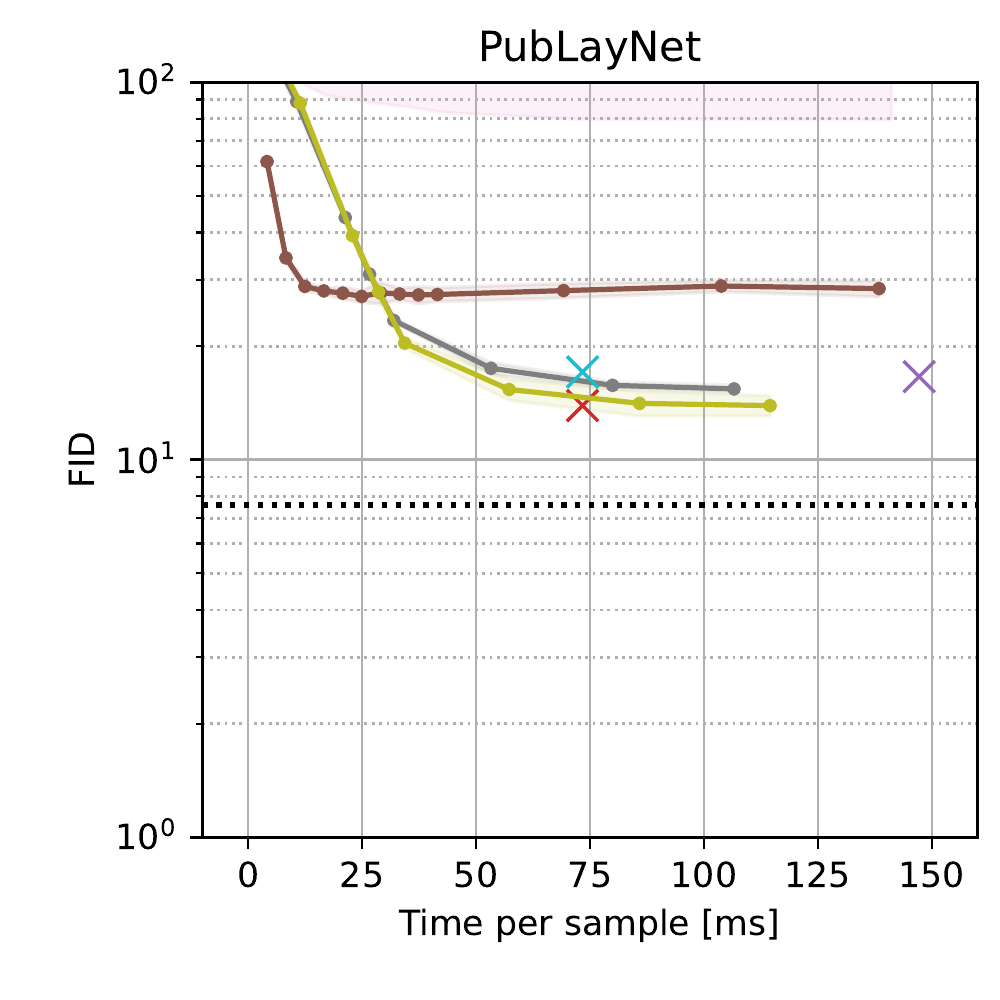} \\
        \multicolumn{2}{c}{
            \adjustbox{trim=0 0 0 0,clip}{
                \includegraphics[width=\linewidth]{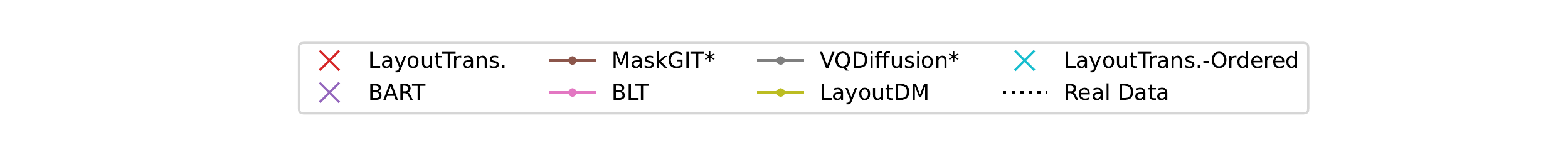}
            }
        }
    \end{tabular}
    \caption{(cont.) Speed-quality trade-off of different models.}
\end{figure*}

\newcommand{\inplength}{0.18}
\newcommand{\inptblheader}{
    & \multirow{2}{*}{\shortstack{Layout-\\Trans.~\cite{gupta2021layout}}} & \multirow{2}{*}{\shortstack{MaskGIT$^\ast$~\cite{chang2022maskgit}}} & \multirow{2}{*}{BLT~\cite{kong2022blt}} & \multirow{2}{*}{BART~\cite{lewis2020bart}} & \multirow{2}{*}{\shortstack{VQDiffusion$^\ast$~\cite{gu2022vector}}} & \multirow{2}{*}{\shortstack{LayoutDM}} \\
    & & & & & &
}
\newcommand{\uctblheader}{
    Real Data & \multirow{2}{*}{\shortstack{Layout-\\Trans.~\cite{gupta2021layout}}} & \multirow{2}{*}{\shortstack{MaskGIT$^\ast$~\cite{chang2022maskgit}}} & \multirow{2}{*}{BLT~\cite{kong2022blt}} & \multirow{2}{*}{BART~\cite{lewis2020bart}} & \multirow{2}{*}{\shortstack{VQDiffusion$^\ast$~\cite{gu2022vector}}} & \multirow{2}{*}{\shortstack{LayoutDM}} \\
    & & & & & &
}
\newcommand{\rftblheader}{
    Input & RUITE~\cite{rahman2021ruite} & \multicolumn{3}{c}{LayoutDM} & Ground Truth \\
}

\newcommand{\inprow}[4]{  %
    \frame{\includegraphics[height=\inplength\hsize]{images/supplementary/comparison/#1/#2/#3_layouttrans_#4.png}} &
    \frame{\includegraphics[height=\inplength\hsize]{images/supplementary/comparison/#1/#2/#3_maskgit_#4.png}} &
    \frame{\includegraphics[height=\inplength\hsize]{images/supplementary/comparison/#1/#2/#3_blt_#4.png}} &
    \frame{\includegraphics[height=\inplength\hsize]{images/supplementary/comparison/#1/#2/#3_bart_#4.png}} &
    \frame{\includegraphics[height=\inplength\hsize]{images/supplementary/comparison/#1/#2/#3_vqdiffusion_#4.png}} &
    \frame{\includegraphics[height=\inplength\hsize]{images/supplementary/comparison/#1/#2/#3_layoutdm_#4.png}}
}
\newcommand{\inprows}[3]{
    Input & & & & & & \\
    \includegraphics[width=0.12\hsize]{images/supplementary/comparison/#1/#2/#3_input.png} & \inprow{#1}{#2}{#3}{0} \\
    Real Data & \inprow{#1}{#2}{#3}{1} \\
    \frame{\includegraphics[width=0.12\hsize]{images/supplementary/comparison/#1/#2/#3_gt.png}} & \inprow{#1}{#2}{#3}{2}
}
\newcommand{\ucrow}[3]{
    \frame{\includegraphics[height=\inplength\hsize]{images/supplementary/comparison/#1/#2/#3_gt.png}} & \inprow{#1}{#2}{#3}{0}
}
\newcommand{\rflength}{0.16}
\newcommand{\rfrow}[3]{
    \frame{\includegraphics[height=\rflength\hsize]{images/supplementary/comparison/#1/#2/#3_input.png}} &
    \frame{\includegraphics[height=\rflength\hsize]{images/supplementary/comparison/#1/#2/#3_ruite_0.png}} &
    \frame{\includegraphics[height=\rflength\hsize]{images/supplementary/comparison/#1/#2/#3_layoutdm_0.png}} &
    \frame{\includegraphics[height=\rflength\hsize]{images/supplementary/comparison/#1/#2/#3_layoutdm_1.png}} &
    \frame{\includegraphics[height=\rflength\hsize]{images/supplementary/comparison/#1/#2/#3_layoutdm_2.png}} &
    \frame{\includegraphics[height=\rflength\hsize]{images/supplementary/comparison/#1/#2/#3_gt.png}}
}

{
    \begin{figure*}[t]
    \centering
    \setlength\tabcolsep{5pt}
    \footnotesize{
        \begin{tabular}{c@{\hspace{3em}}cccccc}
            \inptblheader \\
            \inprows{publaynet}{c}{PMC4284952_00001} \\
             & & & & & & \\
            \inprows{publaynet}{c}{PMC3447363_00002} \\
        \end{tabular}
    }
    \caption{
        Comparison of conditional generation in C$\rightarrow$S+P for PubLayNet.
        We obtain three samples from each model to demonstrate the diversity.
    }
    \label{fig:sup_comparison_c_publaynet}
    \end{figure*}
}

{
    \begin{figure*}[t]
    \centering
    \setlength\tabcolsep{5pt}
    \footnotesize{
        \begin{tabular}{c@{\hspace{3em}}cccccc}
            \inptblheader \\
            \inprows{rico25}{c}{21986} \\
             & & & & & & \\
            \inprows{rico25}{c}{36020} \\
        \end{tabular}
    }
    \caption{
        Comparison of conditional generation in C$\rightarrow$S+P for Rico.
        We obtain three samples from each model to demonstrate the diversity.
    }
    \label{fig:sup_comparison_c_rico}
    \end{figure*}
}

{
    \begin{figure*}[t]
    \centering
    \setlength\tabcolsep{5pt}
    \footnotesize{
        \begin{tabular}{c@{\hspace{3em}}cccccc}
            \inptblheader \\
            \inprows{publaynet}{cwh}{PMC3610354_00004} \\
             & & & & & & \\
            \inprows{publaynet}{cwh}{PMC6081530_00001} \\
        \end{tabular}
    }
    \caption{
        Comparison of conditional generation in C+S$\rightarrow$P for PubLayNet.
        We obtain three samples from each model to demonstrate the diversity.
        Note that the size condition of each element is not shown for limited space.
        Please refer to Real Data for the size.
    }
    \label{fig:sup_comparison_cwh_publaynet}
    \end{figure*}
}

{
    \begin{figure*}[t]
    \centering
    \setlength\tabcolsep{5pt}
    \footnotesize{
        \begin{tabular}{c@{\hspace{3em}}cccccc}
            \inptblheader \\
            \inprows{rico25}{cwh}{6251} \\
             & & & & & & \\
            \inprows{rico25}{cwh}{44501} \\
        \end{tabular}
    }
    \caption{
        Comparison of conditional generation in C+S$\rightarrow$P for Rico.
        We obtain three samples from each model to demonstrate the diversity.
        Note that the size condition of each element is not shown for limited space.
        Please refer to Real Data for the size.
    }
    \label{fig:sup_comparison_cwh_rico}
    \end{figure*}
}

{
    \begin{figure*}[t]
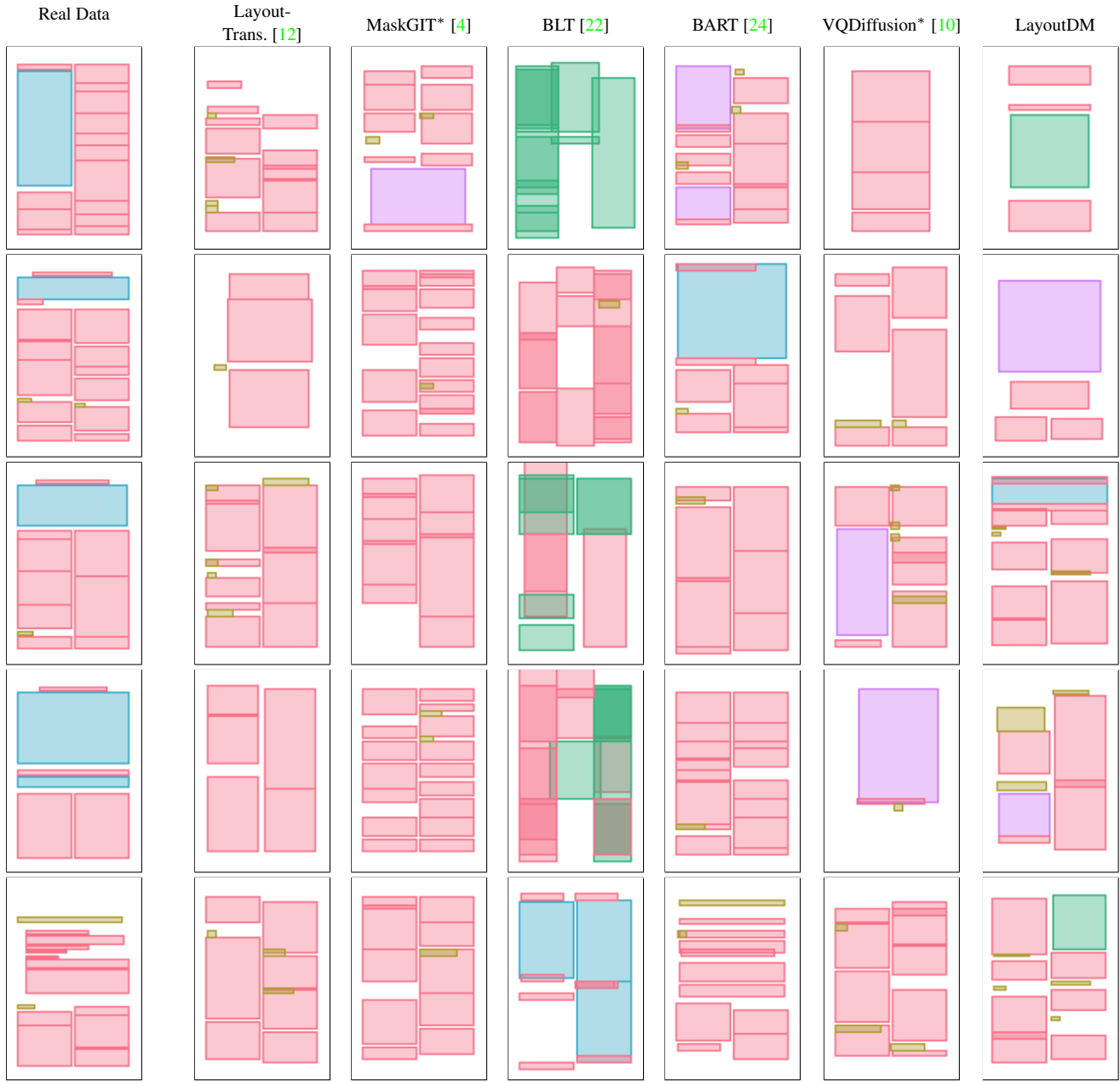

    \centering
    \setlength\tabcolsep{5pt}
    \footnotesize{
        \begin{tabular}{c@{\hspace{3em}}cccccc}
            \uctblheader \\
            \ucrow{publaynet}{unconditional}{PMC3335537_00002} \\
            \ucrow{publaynet}{unconditional}{PMC4971329_00001} \\
            \ucrow{publaynet}{unconditional}{PMC5153480_00002} \\
            \ucrow{publaynet}{unconditional}{PMC4055390_00006} \\
            \ucrow{publaynet}{unconditional}{PMC3503280_00000} \\
        \end{tabular}
    }
    \caption{
        Comparison of unconditional generation for PubLayNet.
        We obtain five samples from each model to demonstrate the diversity.
    }
    \label{fig:sup_comparison_unconditional_publaynet}
    \end{figure*}
}

{
    \begin{figure*}[t]
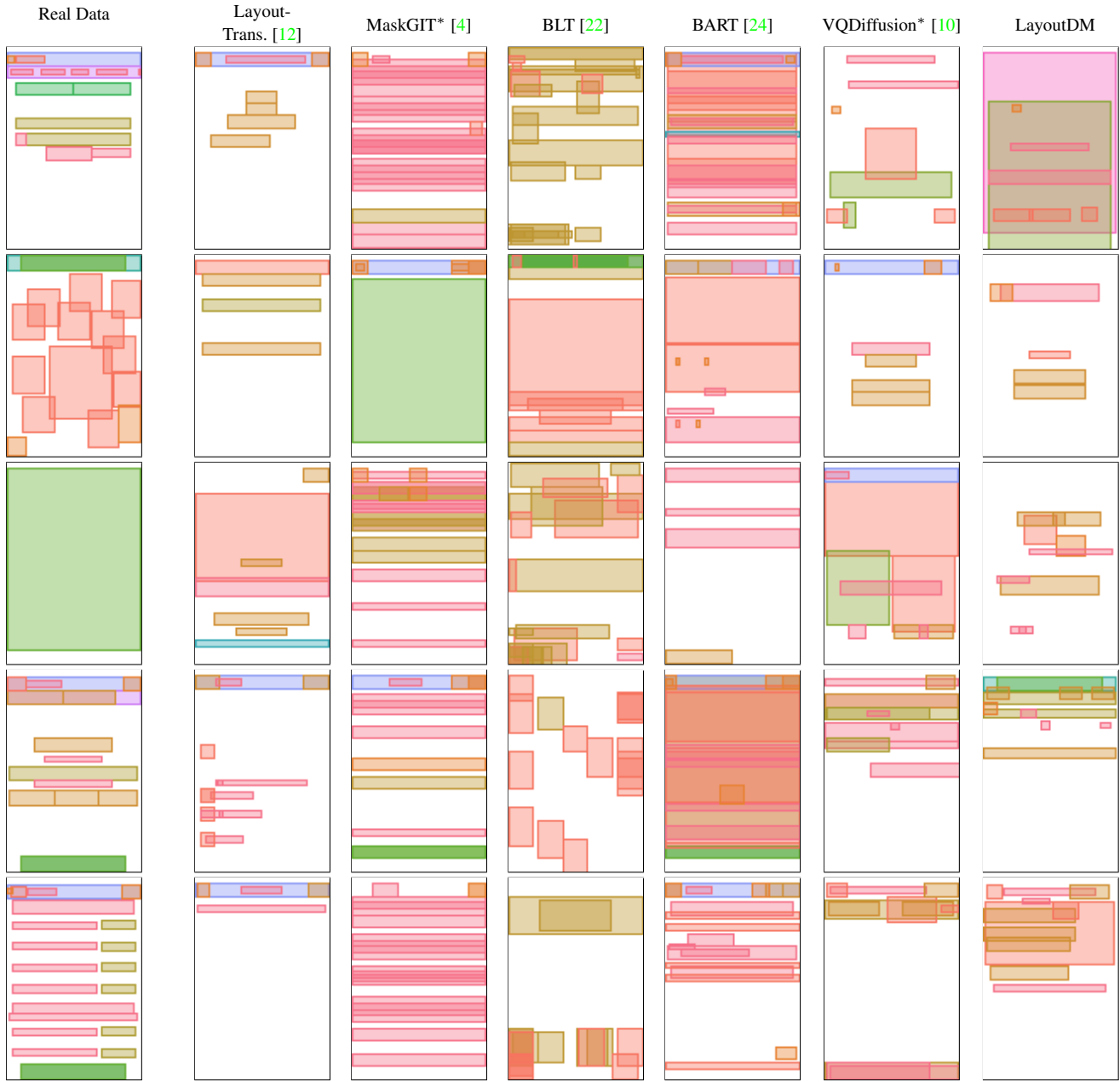

    \centering
    \setlength\tabcolsep{5pt}
    \footnotesize{
        \begin{tabular}{c@{\hspace{3em}}cccccc}
            \uctblheader \\
            \ucrow{rico25}{unconditional}{35360} \\
            \ucrow{rico25}{unconditional}{59406} \\
            \ucrow{rico25}{unconditional}{774} \\
            \ucrow{rico25}{unconditional}{55232} \\
            \ucrow{rico25}{unconditional}{6539} \\
        \end{tabular}
    }
    \caption{
        Comparison of unconditional generation for Rico.
        We obtain five samples from each model to demonstrate the diversity.
    }
    \label{fig:sup_comparison_unconditional_rico}
    \end{figure*}
}

{
    \begin{figure*}[t]
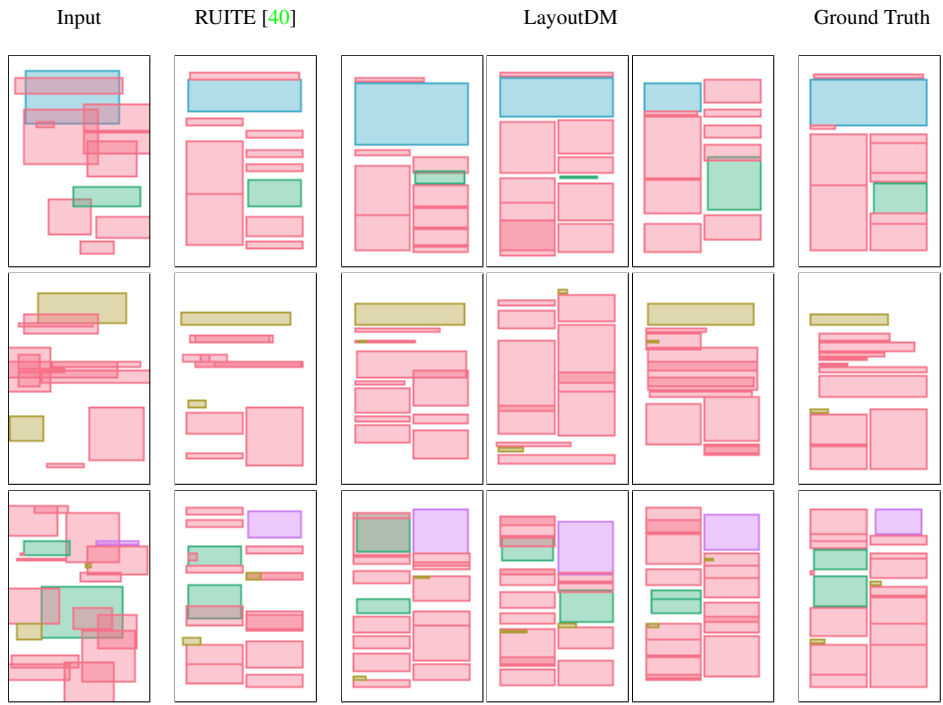

    \centering
    \setlength\tabcolsep{1pt}
    \footnotesize{
        \begin{tabular}{c@{\hspace{10pt}}c@{\hspace{10pt}}ccc@{\hspace{10pt}}c}
            \rftblheader \\
            \rfrow{publaynet}{refinement}{PMC4963558_00004} \\
            \rfrow{publaynet}{refinement}{PMC3255285_00000} \\
            \rfrow{publaynet}{refinement}{PMC3980931_00001} \\
        \end{tabular}
    }
    \caption{
        Comparison of the refinement task for PubLayNet.
        We obtain three samples from LayouytDM to demonstrate the diversity.
    }
    \label{fig:sup_comparison_refinement_publaynet}
    \end{figure*}
}

{
    \begin{figure*}[t]
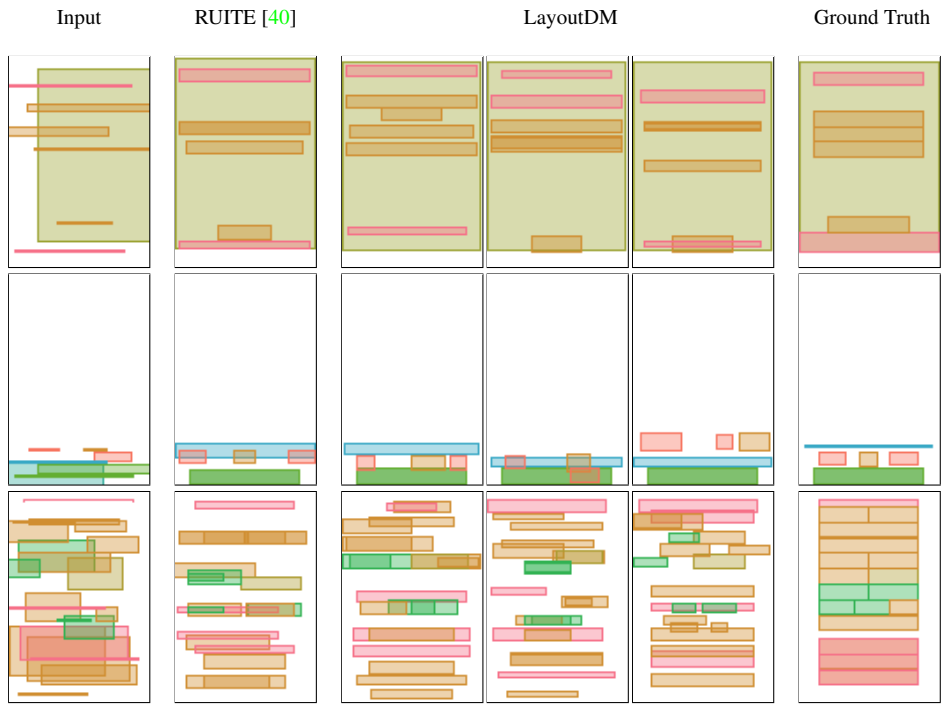

    \centering
    \setlength\tabcolsep{1pt}
    \footnotesize{
        \begin{tabular}{c@{\hspace{10pt}}c@{\hspace{10pt}}ccc@{\hspace{10pt}}c}
            \rftblheader \\
            \rfrow{rico25}{refinement}{31108} \\
            \rfrow{rico25}{refinement}{16902} \\
            \rfrow{rico25}{refinement}{68662} \\
        \end{tabular}
    }
    \caption{
        Comparison of the refinement task for Rico.
        We obtain three samples from LayouytDM to demonstrate the diversity.
    }
    \label{fig:sup_comparison_refinement_rico}
    \end{figure*}
}

\newcommand{\dclength}{0.475}
\newcommand{\dcfig}[1]{
    \adjustbox{trim={.025\width} {.025\height} {.025\width} 0,clip}{
        \includegraphics[width=\dclength\linewidth]{images/supplementary/density_coverage/#1.pdf}
    }
}
\begin{figure*}[t]
    \centering
    \begin{tabular}{cc}
        \multicolumn{2}{c}{C$\rightarrow$S+P} \\
        \dcfig{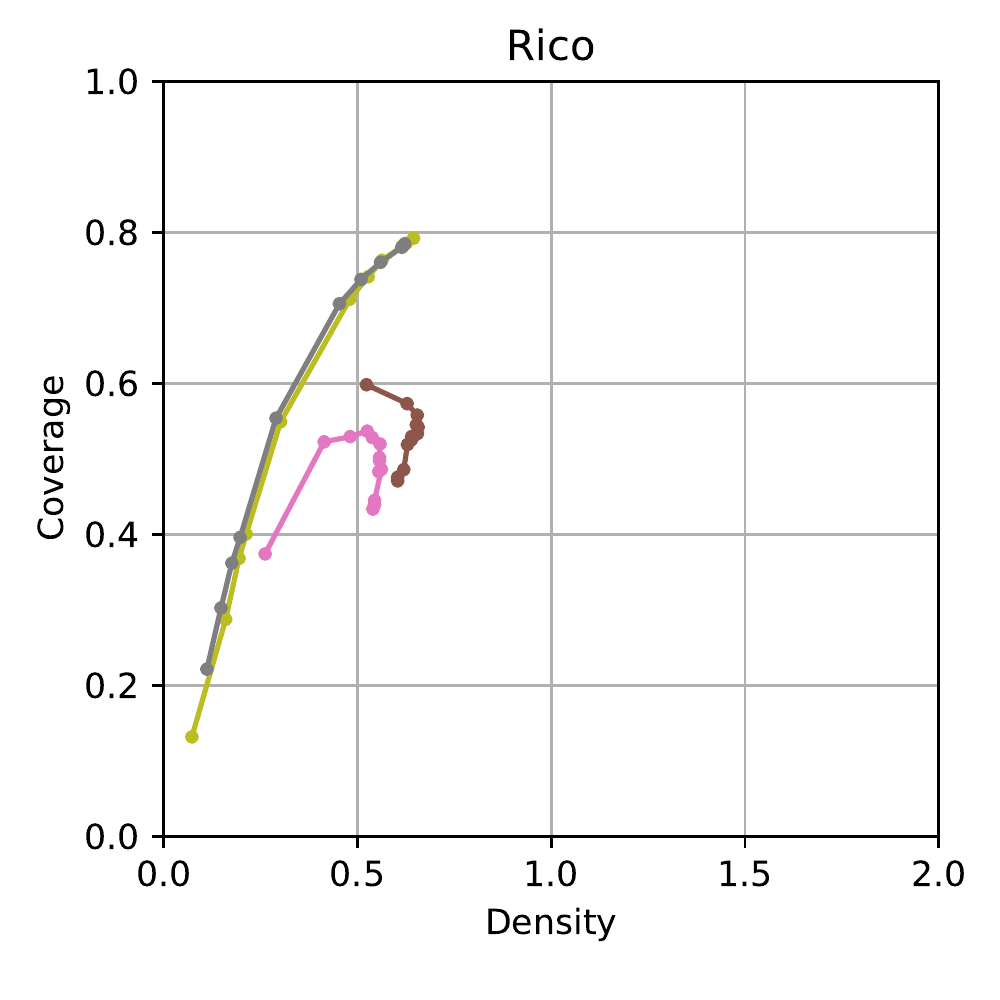} & \dcfig{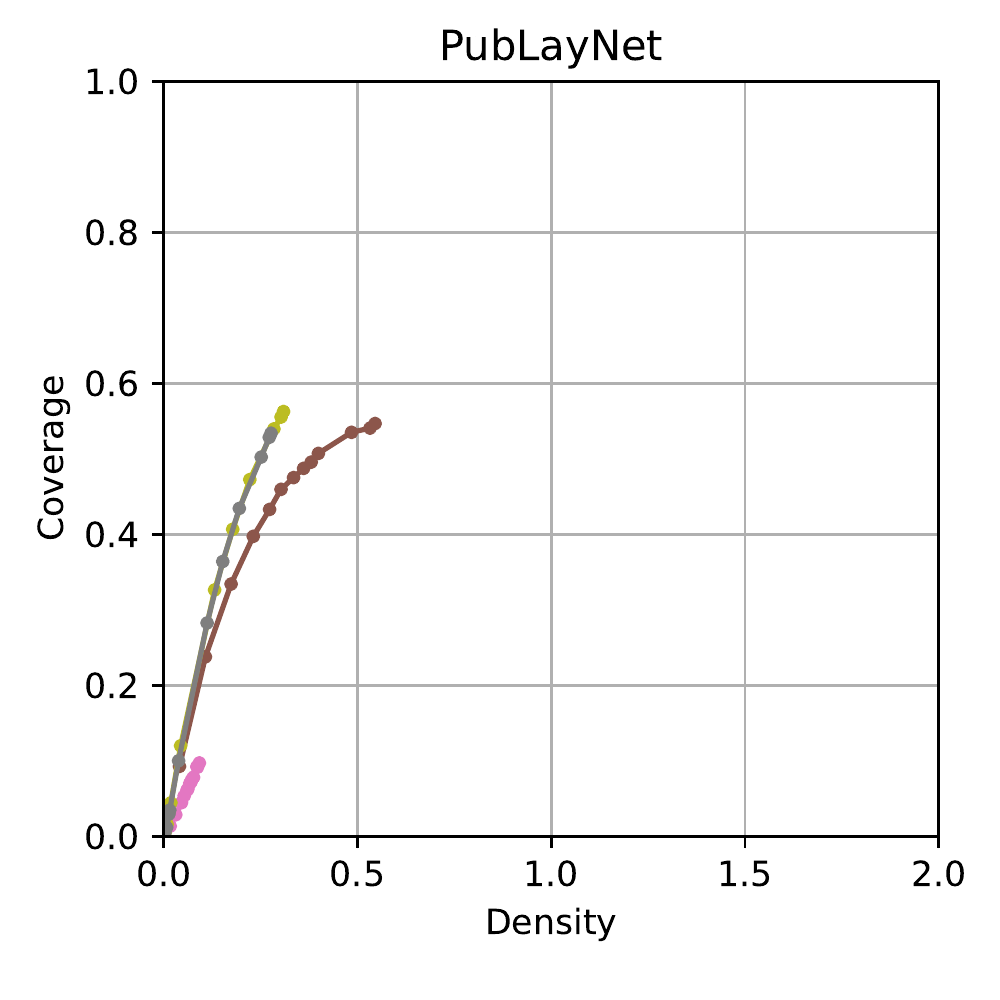} \\
        \multicolumn{2}{c}{
            \adjustbox{trim=0 0 0 0,clip}{
                \includegraphics[width=\linewidth]{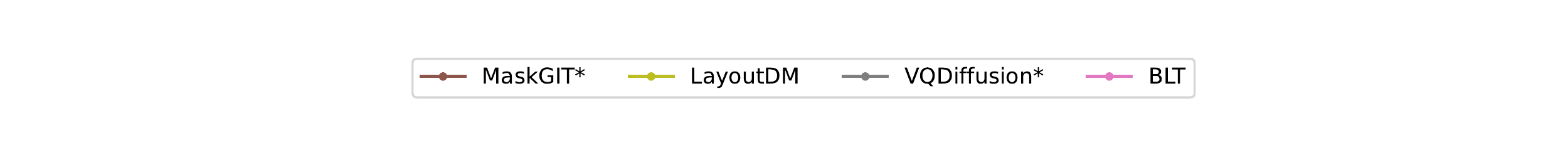}
            }
        } \\
        \multicolumn{2}{c}{C+S$\rightarrow$P} \\
        \dcfig{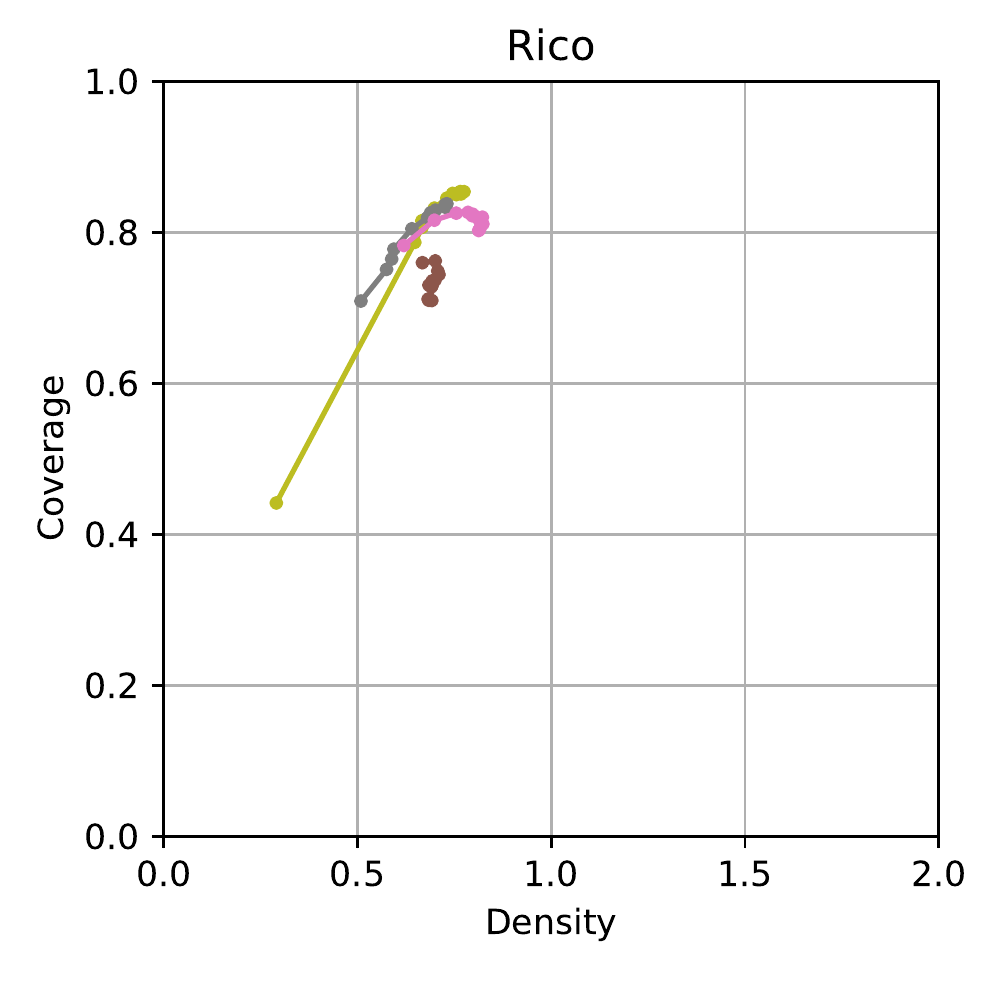} & \dcfig{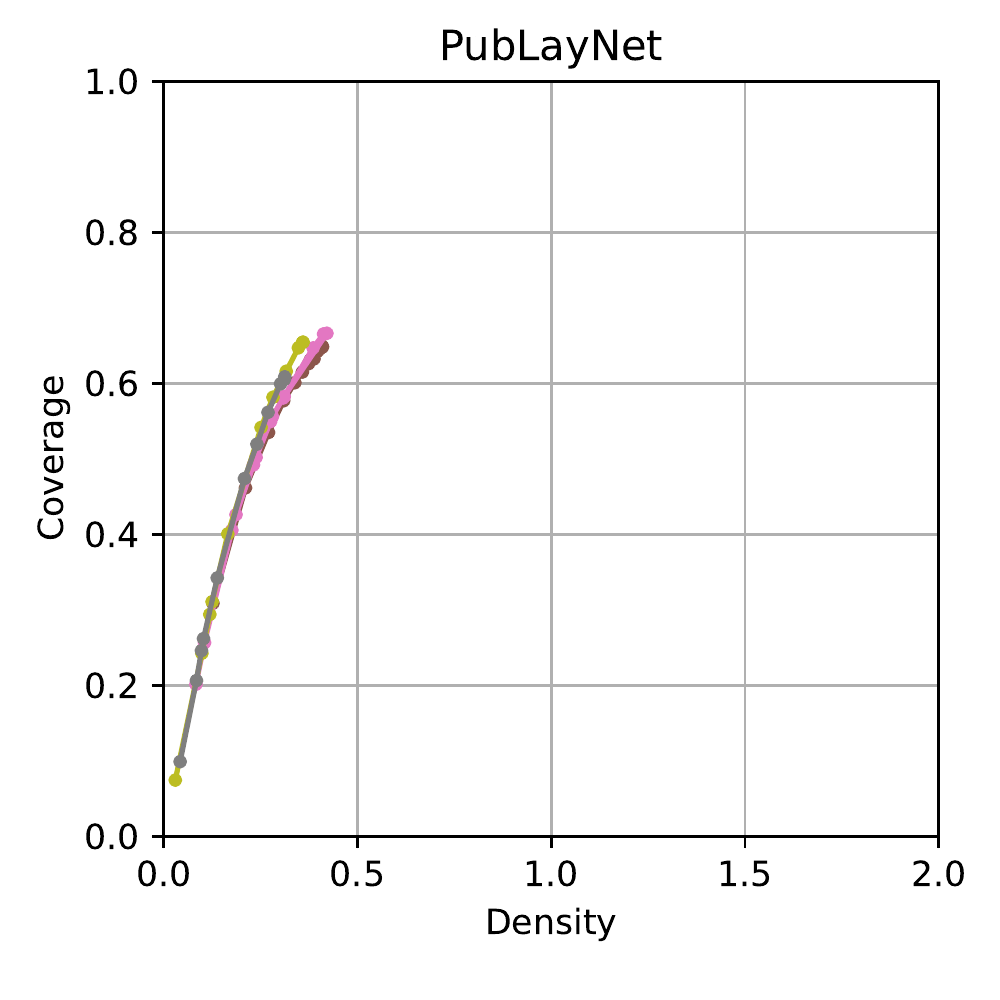} \\
        \multicolumn{2}{c}{
            \adjustbox{trim=0 0 0 0,clip}{
                \includegraphics[width=\linewidth]{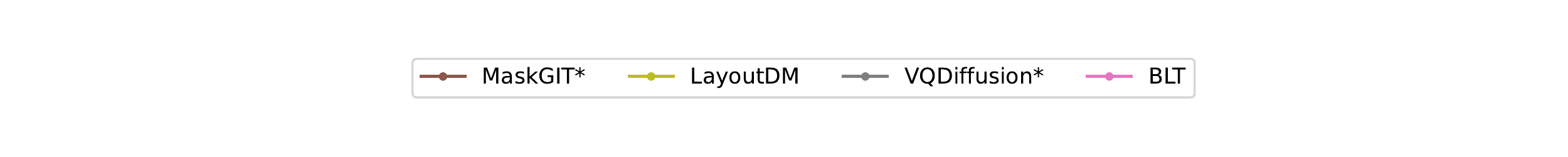}
            }
        }
    \end{tabular}
    \caption{Density-coverage trade-off of different models.}
    \label{fig:sup_tradeoff_density_coverage}
\end{figure*}
\begin{figure*}[t]
    \addtocounter{figure}{-1}
    \centering
    \begin{tabular}{cc}
        \multicolumn{2}{c}{Partial} \\
        \dcfig{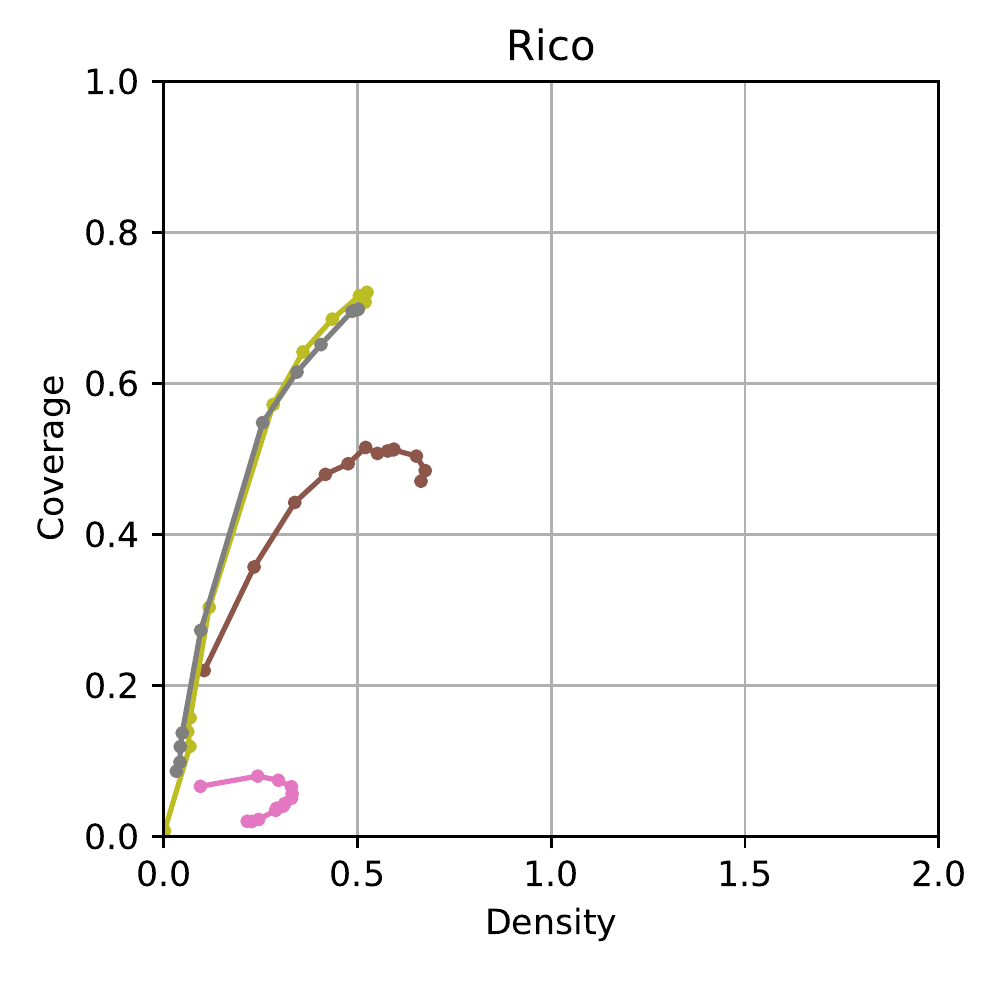} & \dcfig{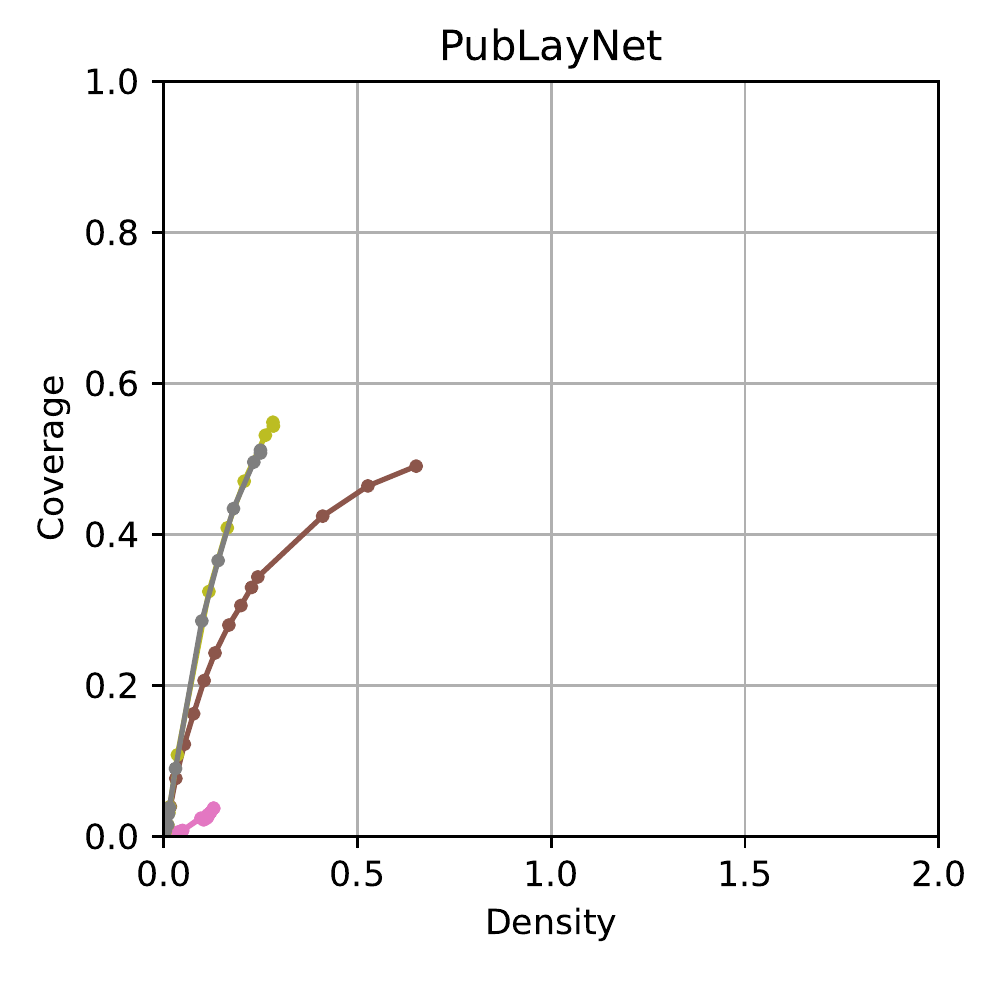} \\
        \multicolumn{2}{c}{
            \adjustbox{trim=0 0 0 0,clip}{
                \includegraphics[width=\linewidth]{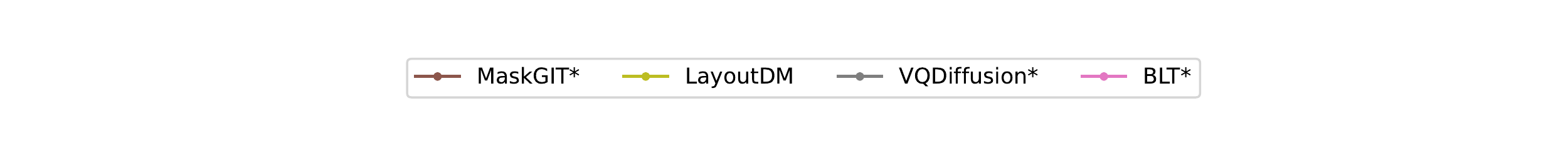}
            }
        } \\
        \multicolumn{2}{c}{Unconditional} \\
        \dcfig{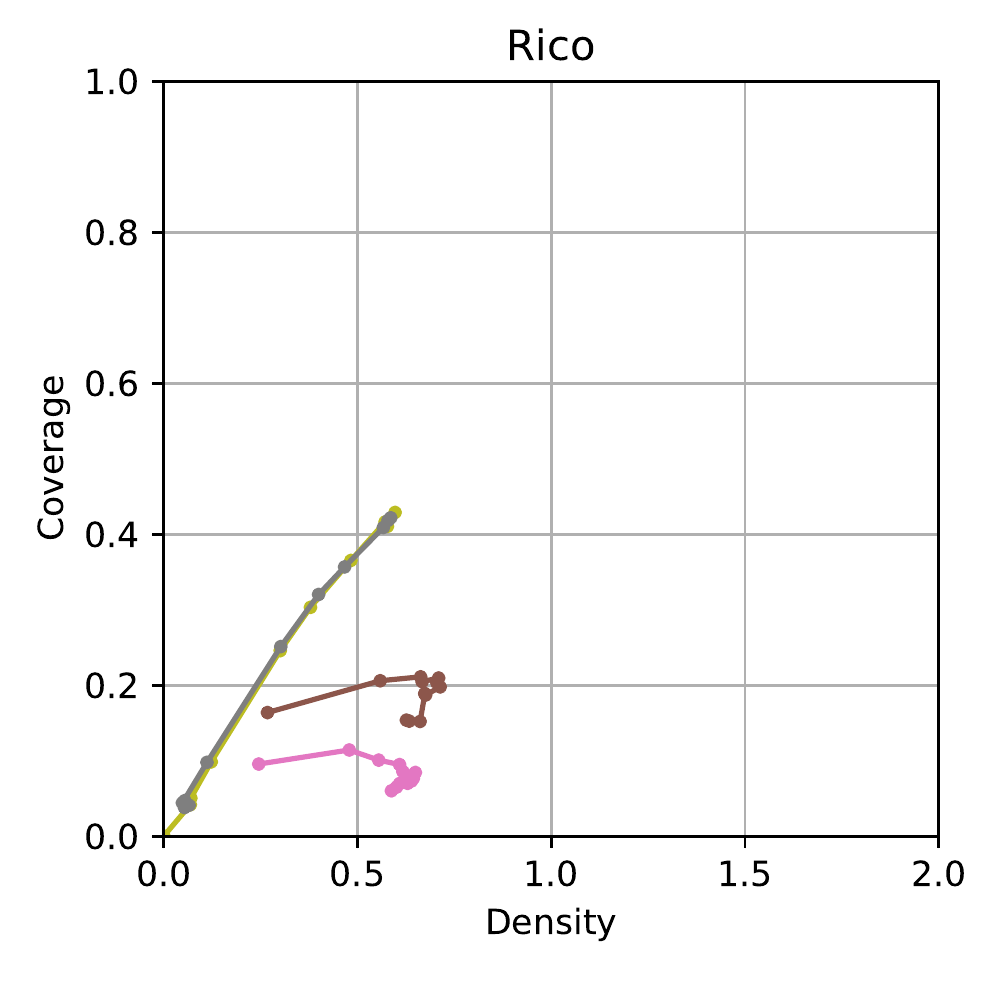} & \dcfig{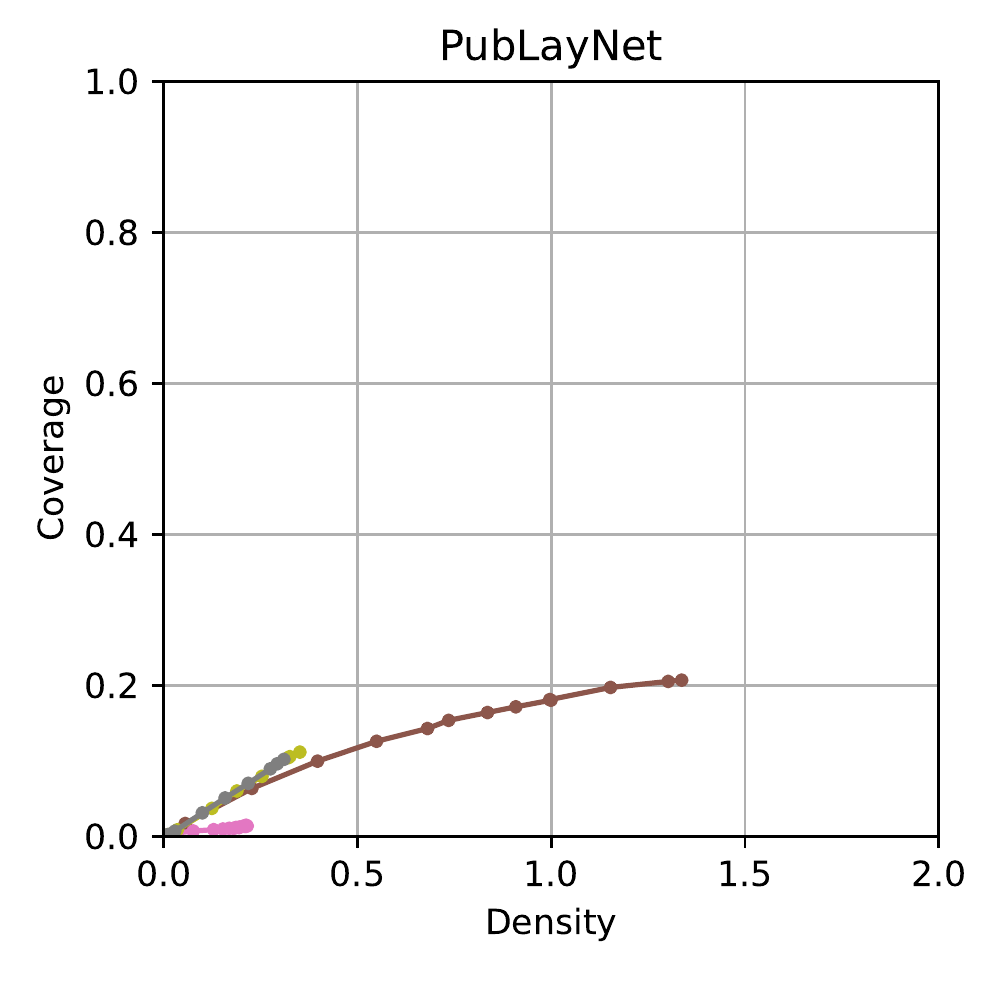} \\
        \multicolumn{2}{c}{
            \adjustbox{trim=0 0 0 0,clip}{
                \includegraphics[width=\linewidth]{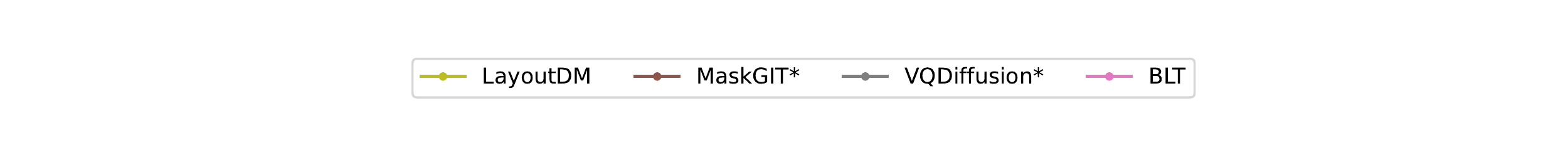}
            }
        }
    \end{tabular}
    \caption{(cont.) Density-coverage trade-off of different models.}
\end{figure*}

\newcommand{\oalength}{0.475}
\newcommand{\oafig}[1]{
    \adjustbox{trim=0 {.025\height} 0 0,clip}{
        \includegraphics[width=\oalength\linewidth]{images/supplementary/overlap_alignment/#1.pdf}
    }
}
\begin{figure*}[t]
    \centering
    \begin{tabular}{cc}
        \multicolumn{2}{c}{C$\rightarrow$S+P} \\
        \oafig{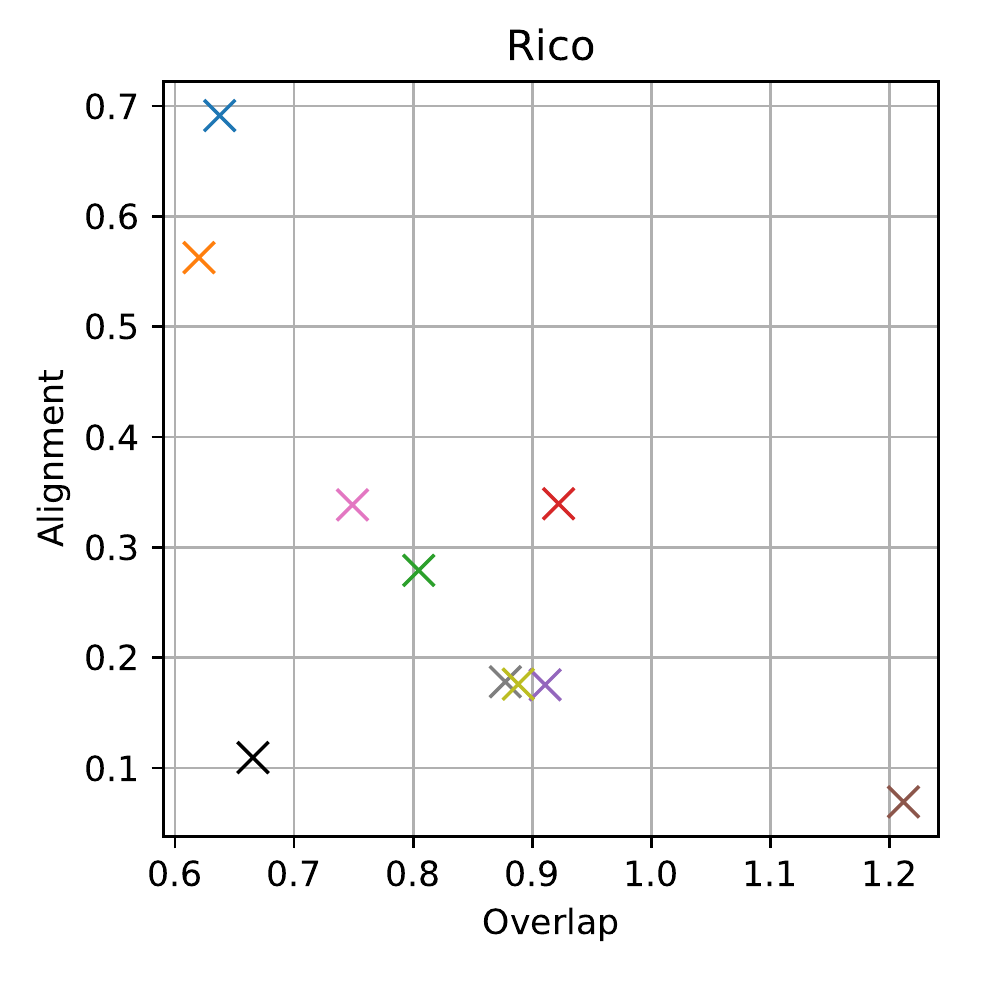} & \oafig{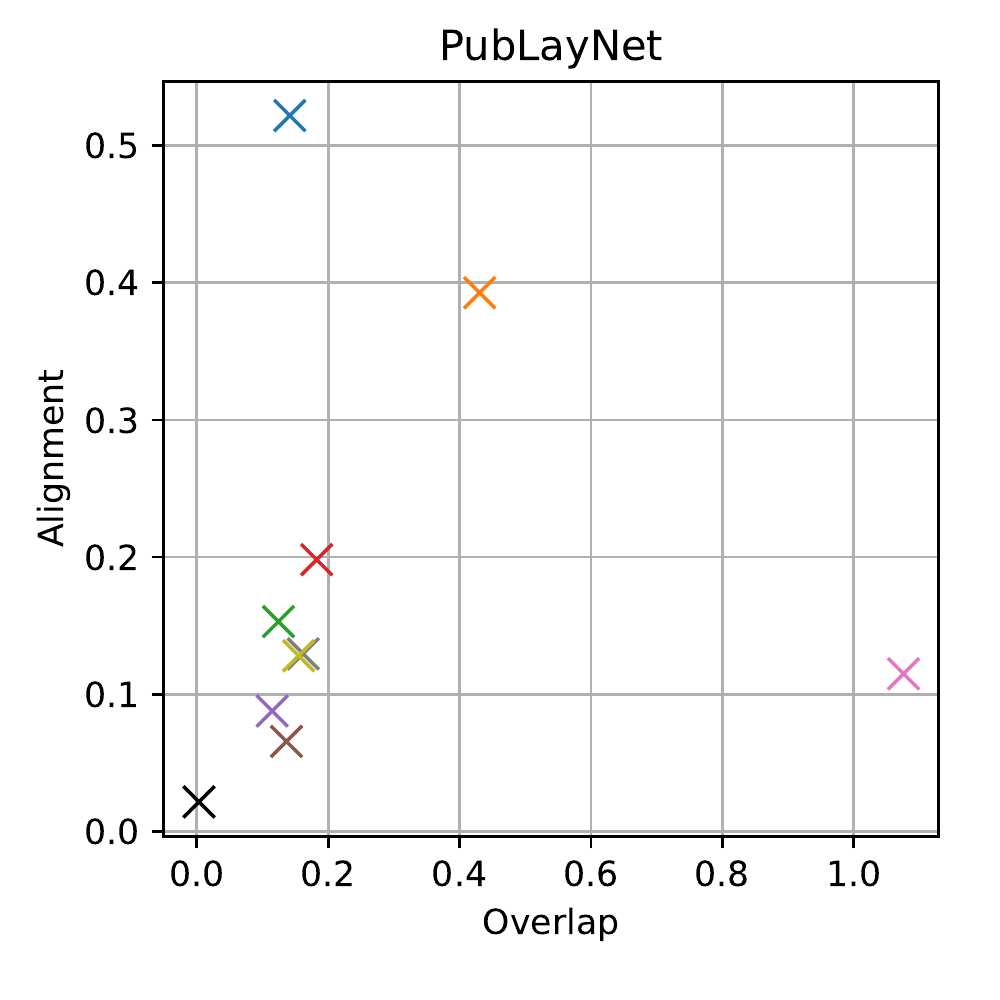} \\
        \multicolumn{2}{c}{
            \adjustbox{trim=0 0 0 0,clip}{
                \includegraphics[width=\linewidth]{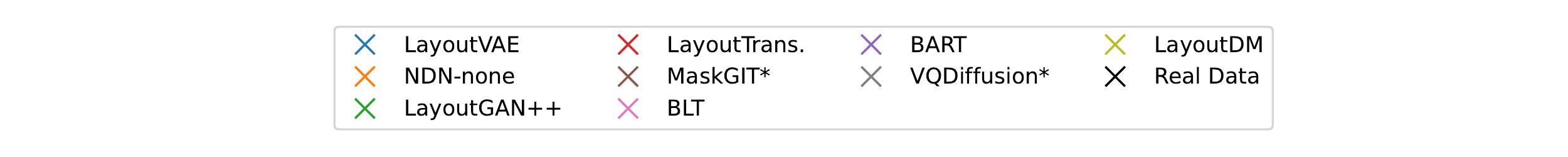}
            }
        } \\
        \multicolumn{2}{c}{C+S$\rightarrow$P} \\
        \oafig{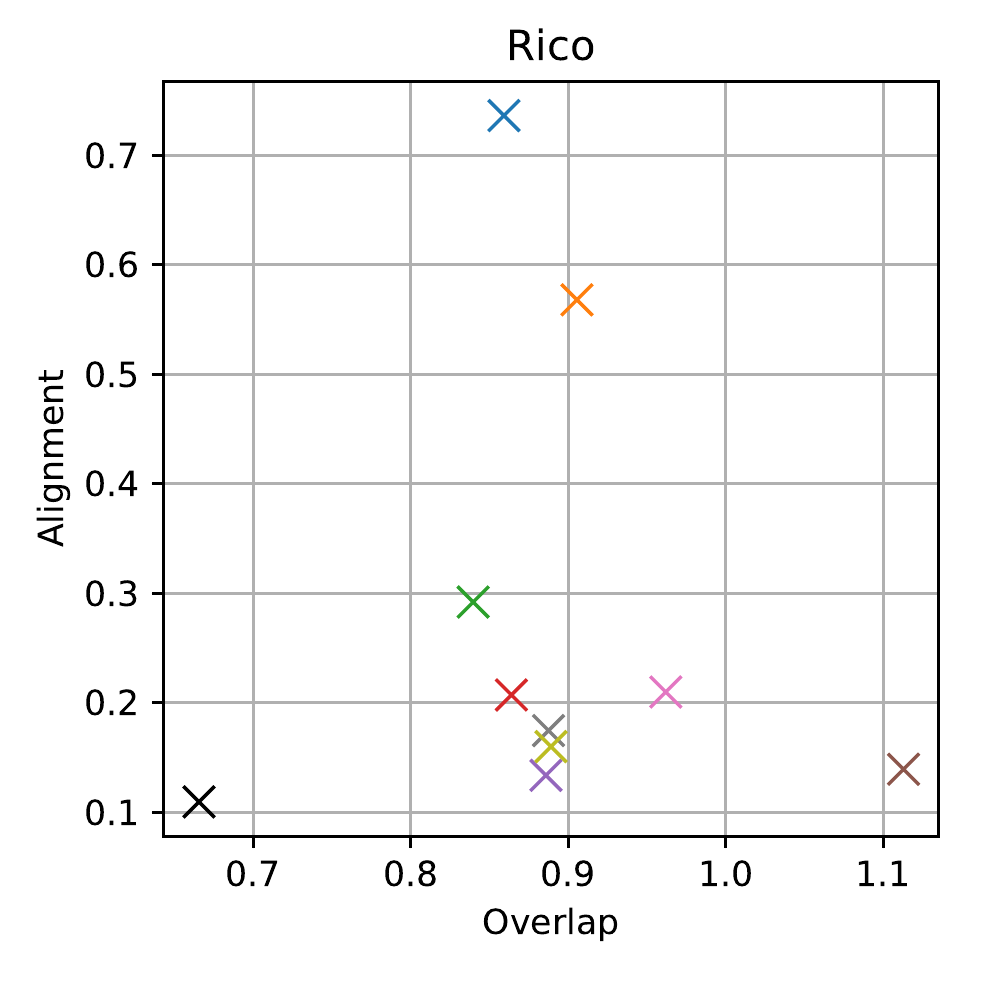} & \oafig{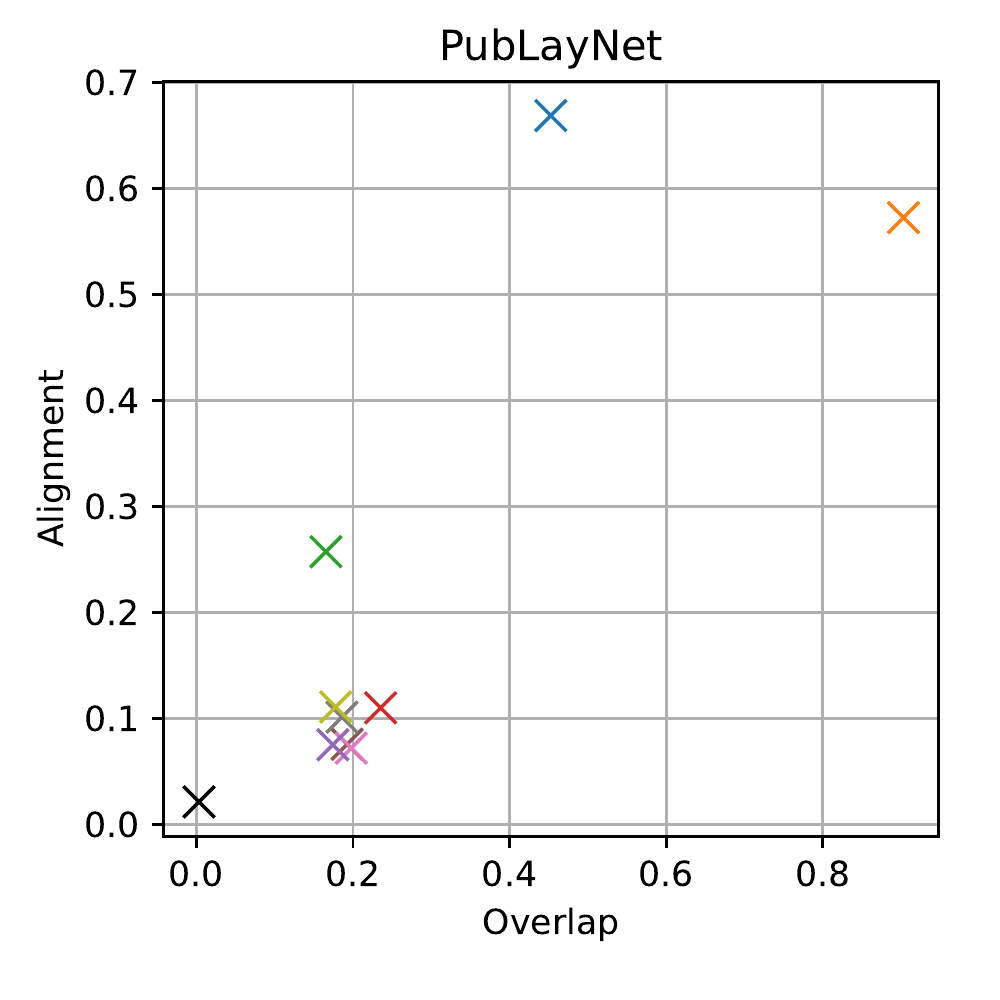} \\
        \multicolumn{2}{c}{
            \adjustbox{trim=0 0 0 0,clip}{
                \includegraphics[width=\linewidth]{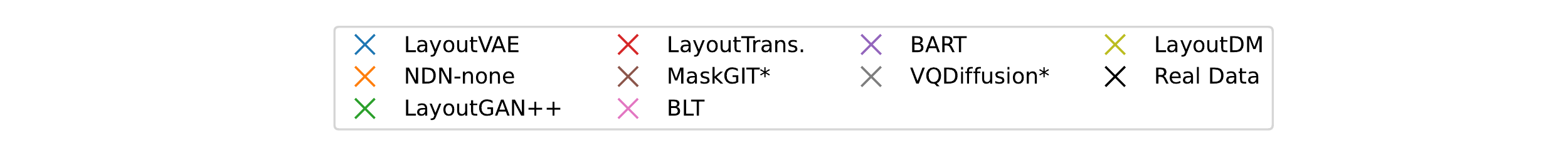}
            }
        } \\
    \end{tabular}
    \caption{Alignment and overlap of different models.}
    \label{fig:sup_overlap_alignment}
\end{figure*}
\begin{figure*}[t]
    \addtocounter{figure}{-1}
    \centering
    \begin{tabular}{cc}
        \multicolumn{2}{c}{Partial} \\
        \oafig{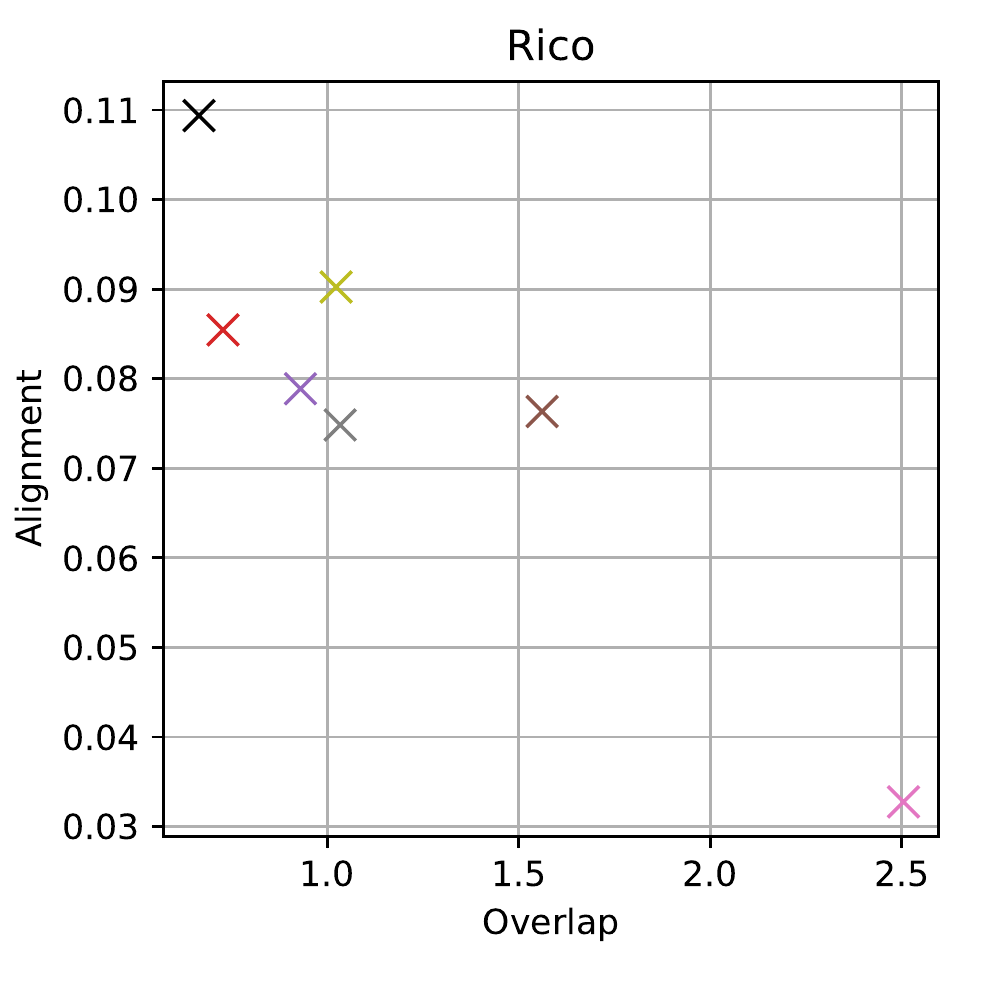} & \oafig{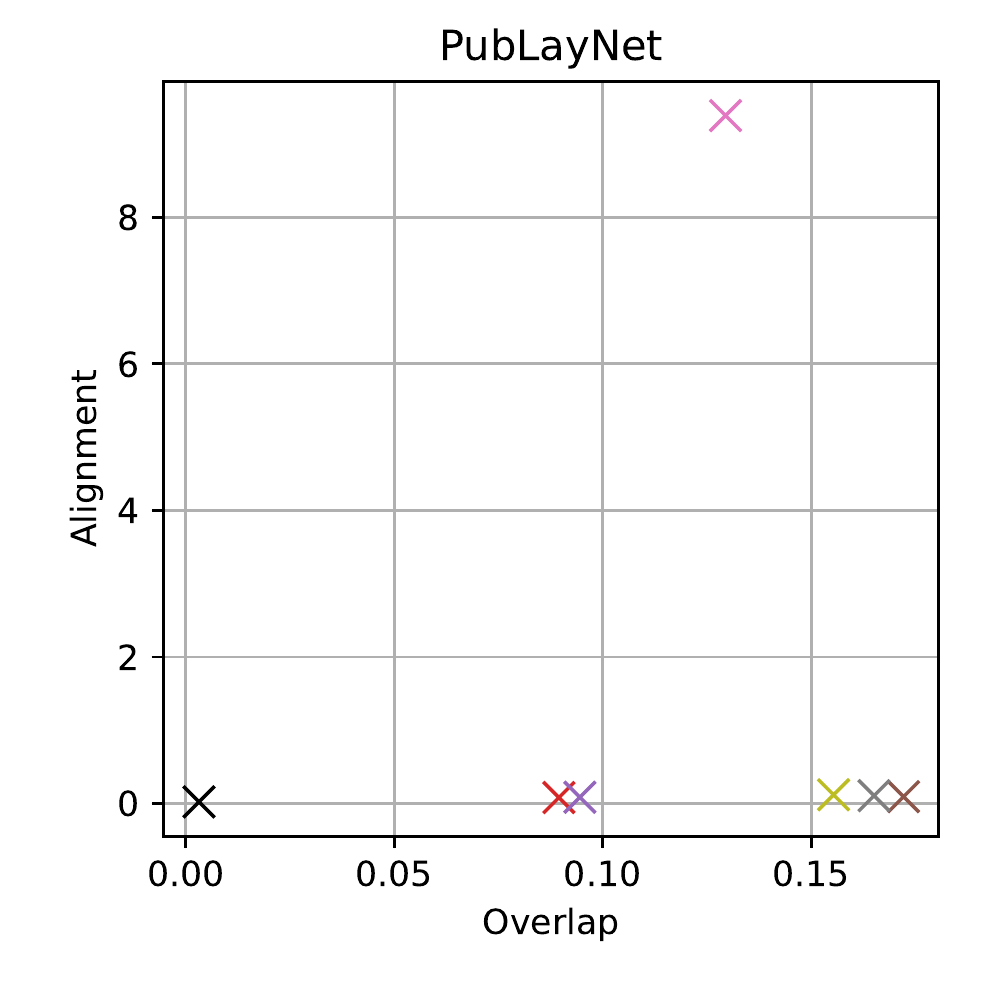} \\
        \multicolumn{2}{c}{
            \adjustbox{trim=0 0 0 0,clip}{
                \includegraphics[width=\linewidth]{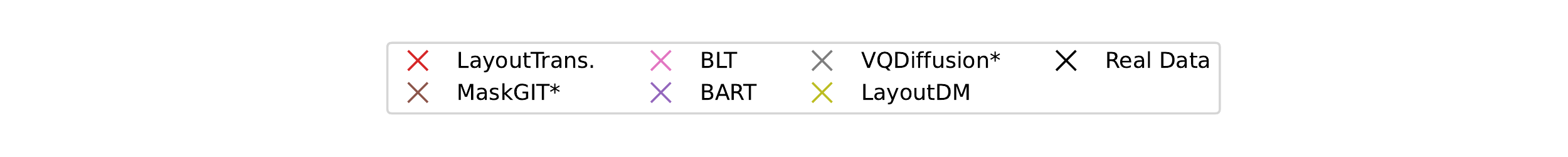}
            }
        } \\
        \multicolumn{2}{c}{Unconditional} \\
        \oafig{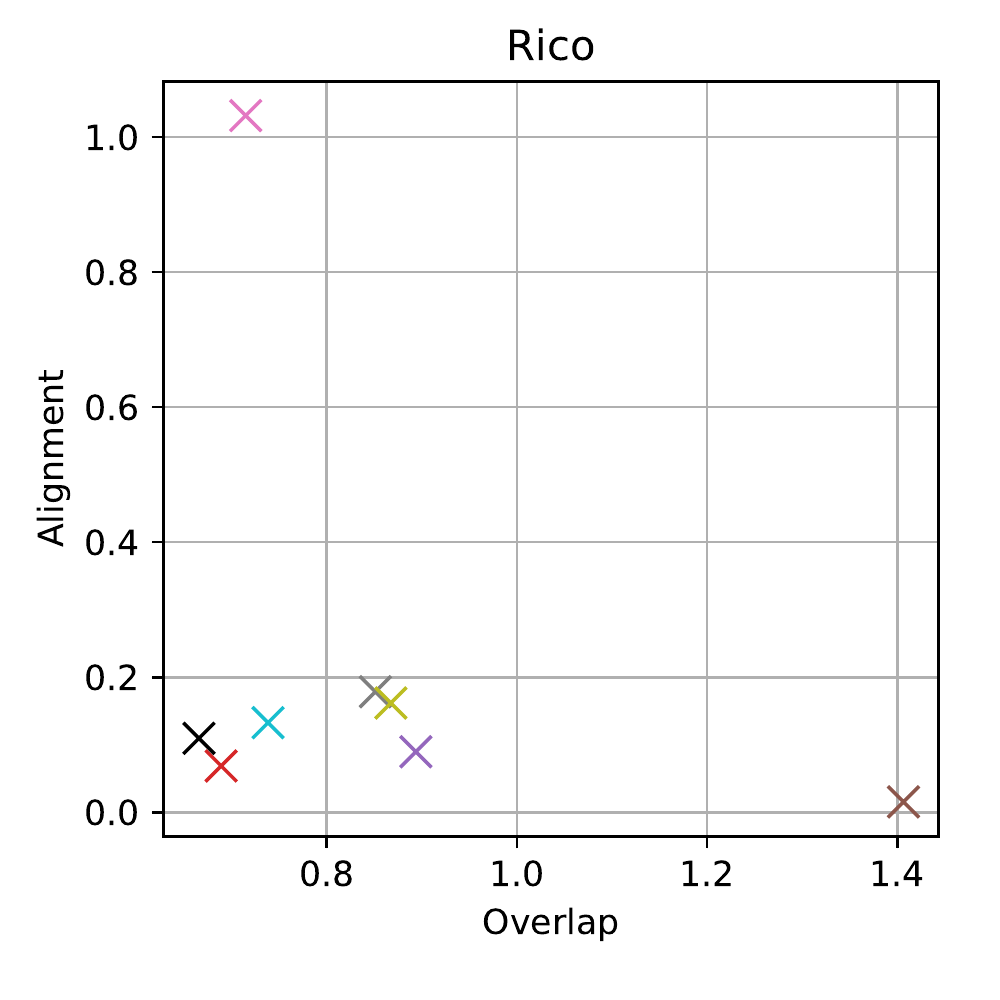} & \oafig{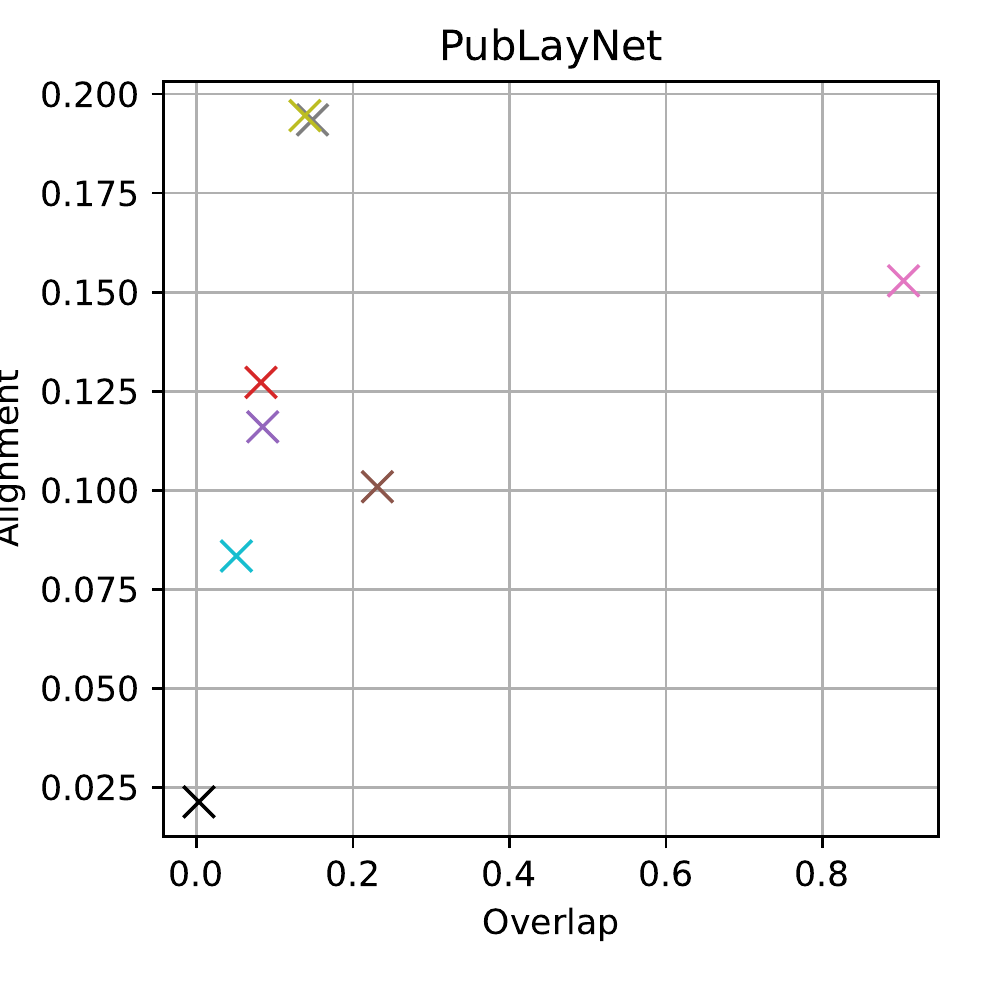} \\
        \multicolumn{2}{c}{
            \adjustbox{trim=0 0 0 0,clip}{
                \includegraphics[width=\linewidth]{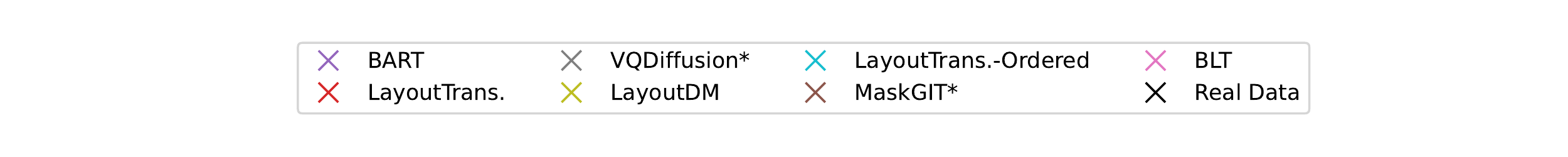}
            }
        } \\
    \end{tabular}
    \caption{(cont.) Alignment and overlap of different models.}
\end{figure*}

\end{document}